\documentclass[preprint,12pt]{elsarticle}

\usepackage{graphicx}
\usepackage{subfig}
\usepackage{amsmath}
\usepackage{amsfonts}
\usepackage[margin=1in]{geometry}
\usepackage{amssymb}
\usepackage{amsbsy}
\usepackage{times}
\usepackage[left]{lineno}
\usepackage{xcolor}
\usepackage{soul}
\usepackage{graphicx,nicefrac}
\usepackage{amsthm}
\usepackage{thmtools}
\usepackage{amsmath,amsfonts,amssymb}
\usepackage{soul}
\usepackage{pgfplots}
\usepackage[figurename=Fig.]{caption}
\usepackage{amsthm}
\usepackage{thmtools}
\usepackage{cases}
\usepackage{algorithm}% <=================http://ctan.org/pkg/algorithms
\usepackage{algpseudocode}% <========== http://ctan.org/pkg/algorithmicx
\usepackage{lineno}
\usepackage{placeins}
\usepackage[normalem]{ulem}
\usepackage{tikz}
\usepackage{hyperref}
\usepackage{scalerel}
\usepackage{url}

\hypersetup{
    colorlinks=true,       % Enable colored links
    linkcolor=blue,        % Color for internal links
    citecolor=blue,        % Color for citation links
    filecolor=blue,        % Color for file links
    urlcolor=blue          % Color for external links
}
\pgfplotsset{compat=newest}
\pgfplotsset{plot coordinates/math parser=false}
\newlength\figureheight
\newlength\figurewidth

\newcommand{\asb}{\textbf}

\newcommand{\inl}[1]{$\textstyle{#1}$}

\newcommand*{\pdot}{\mathbin{\scalerel*{\boldsymbol\odot}{\circ}}}

\graphicspath{/Users/george/Documents/Main_Files/IRZ/Wind field extrapolation}
\definecolor{dgreen}{rgb}{0.0, 0.5, 0.0}

\journal{MSSP}
\begin{document}

\begin{frontmatter}

    \title{ Joint space-time wind field data extrapolation and uncertainty quantification using nonparametric
    Bayesian dictionary learning}

    \author[label1]{George D. Pasparakis}
    
    % \ead{george.pasparakis@irz.uni-hannover.de}

    \author[label2]{Ioannis A. Kougioumtzoglou\corref{cor1}}
    \cortext[cor1]{Corresponding author}
    % \ead{kmd2191@columbia.edu}

    \author[label1]{Michael D. Shields}
    % \ead{iak2115@columbia.edu}

    \address[label1]{Department of Civil and Systems Engineering, Johns Hopkins University, United States of America}
    \address[label2]{Department of Civil Engineering and Engineering Mechanics, Columbia University}
    
    \begin{abstract}
     A methodology is developed, based on nonparametric Bayesian dictionary learning, for joint space-time wind field data extrapolation and estimation of related statistics by relying on limited/incomplete measurements. Specifically, utilizing sparse/incomplete measured data, a time-dependent optimization problem is formulated for determining the expansion coefficients of an associated low-dimensional representation of the stochastic wind field. Compared to an alternative, standard, compressive sampling treatment of the problem, the developed methodology exhibits the following advantages. First, the Bayesian formulation enables also the quantification of the uncertainty in the estimates. Second, the requirement in standard CS-based applications for an a priori selection of the expansion basis is circumvented. Instead, this is done herein in an adaptive manner based on the acquired data. Overall, the methodology exhibits enhanced extrapolation accuracy, even in cases of high-dimensional data of arbitrary form, and of relatively large extrapolation distances. Thus, it can be used, potentially, in a wide range of wind engineering applications where various constraints dictate the use of a limited number of sensors. The efficacy of the methodology is demonstrated by considering two case studies. The first relates to the extrapolation of simulated wind velocity records consistent with a prescribed joint wavenumber-frequency power spectral density in a three-dimensional domain (2D and time). The second pertains to the extrapolation of four-dimensional (3D and time) boundary layer wind tunnel experimental data that exhibit significant spatial variability and non-Gaussian characteristics.
    \end{abstract}

    \begin{keyword}
        Wind data, Stochastic field, Sparse representations, Compressive sampling, Low-rank matrix, Nonparametric Bayesian Dictionary Learning, Boundary Layer Wind Tunnel. 
    \end{keyword}

\end{frontmatter}

\section{Introduction}\label{Introduction}
Long-span structures and high-rise buildings, which are often characterized by light damping and lightweight design, are sensitive to aerodynamic loading. In recent engineering research, investigations of wind-induced structural response are often performed using simulation tools including computational fluid dynamics \cite{jasak2007openfoam,ding2019aerodynamic,kareem2013advanced} methods. However, this approach becomes computationally intractable given the highly nonlinear and unstable behavior of the wind field in bluff body aerodynamics -- especially when coupled with structural response. As a result, extremely dense numerical discretization is required and/or simplifications must be made to solve the related Navier-Stokes equations. 

To circumvent these computational limitations, the wind field can be modeled using stochastic simulation techniques \cite{di1998digital,kareem2008numerical,benowitz2015simulation} in which the wind velocity/pressure is modeled as a stochastic process \cite{roncallo2024fractional}. A rich body of literature has been developed over the past decades to enable the simulation of a wide class of processes: homogeneous/non-homogeneous, stationary/non-stationary, univariate/multivariate, and  Gaussian/non-Gaussian vector process and random fields -- see \cite{deodatis2024spectral,spanos1998monte} for some representative review papers. Recently, an enhanced approach, which employs a stochastic wave representation of the stochastic process, was proposed by Benowitz and Deodatis \cite{benowitz2015simulation}. This enables the simulation of a large number of arbitrarily correlated time-histories in a more efficient manner compared to the standard stochastic vector process representation, which requires costly and numerically unstable matrix decompositions. This approach was later extended to simulate two-dimensional homogeneous wind fields \cite{chen2018simulation} and non-Gaussian stochastic wind fields for long-span structures \cite{zhou2020simulation, vandanapu2024simulation}. Despite recent improvements, stochastic methods are computationally cumbersome as the number of simulation points increases, which is often necessary to account for the considerable spatial variability across the span of the structure. Further, the accuracy is often limited to low-order statistical moments of the target wind field \cite{kareem2020emerging}. 

In practice, wind velocities and pressures are often assessed through Boundary Layer Wind Tunnel (BLWT) testing and field sensor instrumentation. These tools have become the standard for investigating turbulence spectra, mean velocity profiles and wind effects on structures. Although these approaches offer high-fidelity experimental data which are more reliable compared to CFD simulations, BLWT testing is quite expensive, and deploying a large number of sensors (as may be necessary in structural health monitoring applications) can be cost-prohibitive and/or cumbersome. Recent advances in automation and machine learning have improved the accuracy~\cite{catarelli2020automated} and efficiency of BLWT testing~\cite{shields2023active}, but major challenges remain. For example, sensor placement on model structures embedded in BLWT flow is difficult due to the small scale of the model structures and the heterogeneity of the wind pressure field on the surface. 

From both a simulation and an instrumentation perspective, there is significant practical interest in accurately reconstructing, and often extrapolating, joint space-time wind field data and related statistics from the minimum number of observations. For instance, a standard approach for turbulent flow prediction relies on the Reynolds averaged Navier–Stokes (RANS) equations, which, however, can suffer from parametric and/or model-form errors. These are typically calibrated using data-assimilation techniques that minimize the error between the target flow and a partially measured flow field. Representatively, in \cite{foures2014data} the Reynolds stress term in the RANS equations was determined from limited observations using a variational formulation. A related variational data assimilation technique was also used in \cite{brenner2022efficient} to minimize the model error of the eddy viscosity term. Further, the authors in \cite{kontogiannis2024bayesian} used sparse magnetic resonance velocimetry data to solve a Bayesian inverse problem that jointly identified the Navier–Stokes parameters—modeled as Gaussian random variables—and the embedded flow field. In \cite{parish2016paradigm}, a Bayesian inverse problem was solved for determining spatio-temporal correction terms to account for modeling error. The approach was coupled with Gaussian processes to relate the model correction terms with the closure model. Other alternative approaches for flow reconstruction include physics-constrained neural networks \cite{sun2020physics}. Overall, RANS-related methods often focus on average flow properties over relatively small domains, and may require recalibration in response to changes in flow conditions. Moreover, the applicability of neural network-based approaches remains limited due to their lack of robustness when extrapolating to domains associated with limited training data. Thus, there is a need for developing computationally efficient methodologies, endowed with reliable extrapolation capabilities.

In this regard, a methodology is developed in this paper for wind field reconstruction and extrapolation using Bayesian nonparametric statistics and dictionary learning. The proposed methodology builds upon previous work presented in \cite{pasparakis2022wind}, offering three primary contributions. First, it extends the extrapolation capabilities to the 4D domain (3D and time). Second, it circumvents various challenges related to the construction of the expansion basis, which needs to be selected \textit{a priori} in alternative standard compressive sampling (CS) techniques. In contrast, the developed methodology identifies an appropriate expansion basis in an adaptive manner based on a subset of the measured data. Third, it quantifies uncertainties in the estimates that are consistent with the prediction error. The accuracy of the proposed methodology is demonstrated by considering two case studies. The first relates to extrapolating simulated wind velocity records with a prescribed joint wavenumber-frequency power spectral density (PSD) on a 3D domain (2D and time). The second demonstrates the extrapolation of BLWT wind pressure data on a 4D domain (3D and time)~\cite{ojeda2023wind} that exhibit significant spatial variability and non-Gaussian characteristics.

\section{Wind field reconstruction and extrapolation in the joint space-time domain using sparse representations}

Conventional signal processing methods are limited by the Nyquist sampling rate \cite{landau1967sampling}, which states that exact signal reconstruction can be achieved when the sampling frequency is at least twice the highest frequency in the signal. The recently emerged CS theory challenges this paradigm by exploiting the fact that many signals exhibit a sparse structure. A number of theoretical guarantees and related algorithms have been developed to enable signal reconstruction from significantly fewer samples than traditionally required by the Nyquist rate \cite{donoho2006compressed}. In the following section, relevant CS concepts are presented for wind field reconstruction and extrapolation in the joint space–time domain. The interested reader is also directed to the recent review paper \cite{kougioumtzoglou2020sparse} and to references therein for a broad perspective on theoretical concepts and diverse applications of sparse representations and CS approaches in engineering mechanics.

\subsection{Compressive sampling}\label{CS_section}
Consider a collection of wind pressure time histories recorded at \( M = M_x M_z \) spatial points, where \( M_x \) and \( M_z \) represent the number of unique locations in the \( x \)- and \( z \)-directions, respectively. Each point \((x_{i}, z_{i})\) in the 2D spatial domain has a distinct spatial position and has an associated velocity time history \(\mathrm{v}(x_{i}, z_{i}, t)\) recorded at \( N \) time instants. The data are represented in matrix form by  
\begin{equation}\label{y_matrix}
    \asb{Y} = \begin{pmatrix}
        \mathrm{v}(x_1,z_1,t) & \cdots & \mathrm{v}(x_{M_x},z_1,t) \\
        \vdots & \ddots & \vdots \\
        \mathrm{v}(x_{1},z_{M_z},t)  & \cdots & \mathrm{v}(x_{M_x},z_{M_z},t) 
        \end{pmatrix}
\end{equation}
where \( \asb{Y} \in \mathbb{R}^{M_x \times M_z \times N}\). The joint space-time records are reshaped into a single vector \(\mathbf{x}_0 \in \mathbb{R}^{M N \times 1}\) by applying the vectorization operation in the form
\(\mathbf{x}_0 = \text{vec}(\asb{Y}),\)
where \( \mathbf{x}_0 \in \mathbb{R}^{MN} \). Next, it is assumed that this vector can be expanded using an appropriate dictionary of basis functions \(\mathbf{D} \in \mathbb{R}^{M N \times K}\) to yield a system of linear algebraic equations in the form 
\begin{equation}
    \mathbf{x}_0 = \mathbf{D w},
\end{equation} where \(\mathbf{w} \in \mathbb{R}^{K \times 1}\) denotes the coefficient vector and \(K\) is the number of basis functions. The representation is appropriately constructed (e.g., polynomials, Fourier etc.) so that the original vector \(\mathbf{x}_0\) is linearly expanded using a small number, \(s\), of non-zero coefficients where \(s\) is considerably smaller than the number of ambient dimensions, i.e. \(s\ll K\).

In many applications, such as those with limited sensors or simulation points, the records are available only on a subset of locations \(m\) where \(m < M\). In this case, the goal is to extrapolate the missing records to the \(M-m\) locations where data are missing. Here, the observation vector is written as 
\begin{equation}\label{CS}
    \mathbf{x} = \mathbf{\Phi} \mathbf{x}_0 = \mathbf{A w}
\end{equation}
where \(\mathbf{A} = \mathbf{\Phi D}\) is called the sampling matrix and \(\mathbf{\Phi}\) is an \(mN \times M N\) binary measurement matrix (also known as the CS matrix \cite{candes2006compressive}) which deletes the rows of \(\mathbf{D}\) corresponding to the missing locations. Thus, \( \mathbf{x} \in \mathbb{R}^{mN} \). Under the assumption of sparsity and other important properties from CS theory \cite{elad2010sparse}, the coefficient vector \(\mathbf{w}\) can be uniquely determined despite the system of equations in Eq.~\eqref{CS} being underdetermined. This is achieved by solving a constrained optimization problem of the general form
\begin{equation}\label{p_norm_min_const}
    \min_{\mathbf{w}}\mid\mid \mathbf{w} \mid\mid_p \textrm{ subject to }  \mid\mid \mathbf{x} -  \mathbf{A w} \mid\mid_2 \leq \epsilon
\end{equation}
where \(\mid\mid \mathbf{w} \mid\mid_p=\left(\sum_{j=1}^k |w_j|^p \right)^{\frac{1}{p}}\) denotes the \inl{\ell_p}-norm. To facilitate its numerical solution, Eq.~\eqref{p_norm_min_const} can be equivalently written as an unconstrained optimization problem in the form 
\begin{equation}\label{p_norm_min_unconst}
    \min_{\mathbf{w}}\mid\mid \mathbf{x} -  \mathbf{A w} \mid\mid_2 + \ \lambda\mid\mid \mathbf{w} \mid\mid_p,
\end{equation}
where \inl{\lambda} is a penalization factor. Once the sparse coefficient vector \(\mathbf{w}\) is determined, it can be used to construct \(\mathbf{x}_0\) from Eq.~\eqref{CS}. Various theoretical guarantees and related algorithms can be employed depending on the choice of the \( p \)-norm penalty \cite{rish2014sparse}. For instance, the theoretically optimal solution is determined by minimizing the \(\ell_0\)-norm of the coefficient vector \(\mathbf{w}\). However, this requires a combinatorial search over all possible solutions which is an NP-hard problem (NP stands for non-deterministic polynomial time). To address this, there exist approximate schemes, such as the Orthogonal Matching Pursuit (OMP) algorithm \cite{pati1993orthogonal} which build the solution in an iterative manner. Alternatively, \(\ell_1\)-norm minimization schemes replace the original \(\ell_0\) norm minimization with a convex surrogate assuming that the matrix \(\mathbf{A}\) satisfies the restricted isometry property (RIP) \cite{candes2006stable}. This leads to a convex optimization problem for which minimization procedures are readily available \cite{boyd2004convex}. 

Traditional CS schemes rely on prior knowledge of the expansion basis \(\mathbf{D}\), which presents certain limitations. Although many popular dictionaries (e.g., discrete cosine transform and wavelets) exhibit satisfactory performance for one-dimensional signals and two-dimensional images, they may not be well suited to represent arbitrary and/or high-dimensional data \cite{rubinstein2010dictionaries}. Furthermore, even if such a basis theoretically exists, its construction can be computationally daunting. For example, for 2D wind field reconstruction \cite{pasparakis2022wind}, the sampling matrix \(\asb{A}\) can be expressed using the spectral representation method \cite{chen2018simulation}. Following Pasparakis et al. \cite{pasparakis2022wind}, matrix \(\asb{D}\) is formed as the tensor product of basis functions spanning the frequency (\(\omega\)) domain and the wavenumber domains (\(\kappa^x\) and \(\kappa^z\)) in the \(x\) and \(z\) direction, respectively. This is given as 
\begin{equation}
    \mathbf{B}_t = \left[\cos(\omega_0 t), \sin(\omega_1 t), \cdots, \sin(\omega_{\frac{N}{2}-1} t), \cos(\omega_{\frac{N}{2}} t) \right]
\end{equation}
where \(\omega_\ell = 2\pi \ell / N\) and \(\ell = 0, \cdots, N/2\),
\begin{equation}
    \mathbf{B}_x = \left[\cos(\kappa^x_{0} x), \sin(\kappa^x_{1} x), \cdots, \sin(\kappa^x_{\frac{M_x}{2}-1} x), \cos(\kappa^x_{\frac{M_x}{2}} x) \right]
\end{equation}
with \(\kappa^x_{m} = 2\pi m / M_x\) and \(m = 0, \cdots, M_x/2\),
\begin{equation}
    \mathbf{B}_z = \left[\cos(\kappa^z_{0} z), \sin(\kappa^z_{1} z), \cdots, \sin(\kappa^z_{\frac{M_z}{2}-1} z), \cos(\kappa^z_{\frac{M_z}{2}} z) \right]
\end{equation}
where \(\kappa^z_{m} = 2\pi m / M_z\) and \(m = 0, \cdots, M_z/2\), respectively. The combined basis \(\mathbf{D}\) is then formed by taking the tensor product of \(\mathbf{B}_t\), \(\mathbf{B}_x\), and \(\mathbf{B}_z\)
\begin{equation}\label{D_tensor}
    \mathbf{D} = \mathbf{B}_t \otimes \mathbf{B}_x \otimes \mathbf{B}_z
\end{equation}
It is worth noting that overcomplete dictionaries can be used to construct \(\mathbf{B}_t\), \(\mathbf{B}_x\), and \(\mathbf{B}_z\), allowing for enhanced resolution in their respective domains \cite{chen2001atomic}.

In recent years, sparse representations have been widely used in the context of fluid mechanics. For example, a library of flow snapshots together with a CS approach was employed in \cite{callaham2019robust} to reconstruct the target flow field. A sparse regression procedure was employed in \cite{loiseau2018sparse}
to identify reduced-order dynamics from particle image velocimetry (PIV) data. Further, physics-informed CS approaches were proposed in \cite{kontogiannis2022physics,lu2020physics}
to solve inverse Navier–Stokes boundary value problems. Despite their success, sparsity-based methods are limited in modeling high-dimensional and/or complex data. For instance, the resulting Fourier matrix \(\mathbf{D} \in \mathbb{R}^{N M_x M_z \times N M_x M_z}\) in Eq.~\eqref{D_tensor} requires an increasingly large memory size that grows rapidly with increasing dimensions and, thus, the associated optimization in Eq.~\eqref{p_norm_min_unconst} is cost prohibitive. Moreover, the incomplete nature of the time history data introduces consecutive missing segments in the sampling matrix, which may violate the RIP condition. Lastly, the underlying harmonic expansion assumes a stationary Gaussian wind field process, which may not be valid in practical applications \cite{kareem2020emerging}. To address these limitations, the following sections propose alternative ``basis-free" methodologies for wind field reconstruction and extrapolation in two and three spatial dimensions.

\subsection{Low-rank matrix recovery}\label{Low_rank_section}
In this section, a computationally tractable approach to extrapolate high-dimensional wind data is proposed using a matrix completion algorithm. This family of methods relies on the fact that, in many cases, an appropriate rearrangement of the data yields a low-rank matrix, where the low-rank property can be interpreted as the generalization of sparsity for matrices (see, e.g., \cite{meyer2000matrix,friedland2018nuclear}). Thus, in analogy to sparse vectors, a partially observed matrix can be reconstructed by recovering the minimum-rank matrix among all possible matrices that are consistent with the observed entries. However, rank minimization is NP-hard and, therefore, intractable in practical applications. Nevertheless, motivated by the success of the \(\ell_1\) norm minimization, a convex surrogate for the rank can be adopted. Observing that the rank of a matrix is given as the number of its non-zero singular values, a natural extension is to replace the rank function by the nuclear norm, which is the sum of its singular values. For a matrix \(\asb{Y} \in R^{n_1  \times  n_2}\) the nuclear norm is given by 
\begin{equation}
    \|\asb{Y}\|_* = \sum_{i=1}^{\min\{n_1, n_2\}} \sigma_i(\mathbf{\asb{Y}})
\end{equation}
where \(\|\cdot\|_{*}\) denotes the nuclear norm and \(\sigma_i(\asb{Y})\) is the \(i\)\textit{th} singular value of \(\asb{Y}\). Early results in matrix completion \cite{candes2009exact} demonstrate that exact reconstruction is guaranteed when \( m \geq C n^{5/6} r\log n \), in which \(m\) is the number of observations, \(C\) is a positive constant, \(n=\min\{n_1, n_2\}\), and \(r\) is the rank of the matrix. The aforementioned bound has been subsequently improved and specific relationships between \(n\), \(r\), and \(l\) can be found in \cite{candes2009exact,candes2010power}. Overall, the required number of observations is considerably smaller than the total number of entries \(n_1 n_2\), whereas the sublinear scaling bound yields a decreasing fraction of observations as the matrix dimensions increase. Also, a smaller rank requires fewer observed matrix entries for successful matrix completion.

In view of the above and anticipating that the wind field is sufficiently correlated in the spatial domain, a matrix reshape of the wind data is adopted in conjunction with nuclear norm minimization to recover the missing matrix entries. This is formally expressed as 
\begin{equation}
    \begin{array}{ll}
        \mathrm{min} & \|{\asb{Y}^{(i)}}\|_{*}\\
        \text{subject to } & \asb{Y}_{k,l}^{(i)}=\asb{M}_{k,l}^{(i)}, \quad(k, l) \in \Omega,
        \end{array}
        \label{nuclear_min}
\end{equation}
where \(\asb{Y}^{(i)} = \asb{Y}(t_i) \in \mathbb{R}^{M_x \times M_z}\) (see Eq.~\eqref{y_matrix}), \(r<<M_x \times M_z\) and \(\Omega\) is the index set of observed entries given by matrix \(\asb{M}\). Minimizing Eq.~\eqref{nuclear_min} leads to a convex optimization problem, which can be treated by a variety of numerical schemes; see, e.g., \cite{cai2010singular, lin2010augmented, kougioumtzoglou2020sparse} and the references therein. In this paper, the Augmented Lagrangian Method (ALM) is used \cite{lin2010augmented} so that the original problem in Eq.~\eqref{nuclear_min} is reformulated as
\begin{equation}
    \begin{array}{ll}
        \mathrm{min} & \|{\asb{Y}^{(i)}}\|_{*}\\
        \text{subject to } & \asb{Y}^{(i)} + \asb{E} = \asb{M}^{(i)}, \ \mathcal{P}_{\bar{\Omega}}(\asb{E}) = 0
    \end{array}
    \label{nuclear_min_E}
\end{equation}
where \(\asb{E}\) is a discrepancy matrix which accounts for the unknown entries of \({\asb{Y}^{(i)}}\), and \(\mathcal{P}_{\bar{\Omega}}(\cdot)\) is a linear operator that sets the observed entries outside of \(\Omega\) (i.e., in \(\bar{\Omega}\)) to zero. This approach can be viewed as a modified version of the Robust PCA \cite{wright2009robust}, which yields an optimal estimate for \(\asb{Y}^{(i)}\) for arbitrarily large values of the corruption matrix \(\asb{E}\) as long as \(\asb{E}\) is sufficiently sparse relative to the rank of \(\asb{Y}^{(i)}\). To facilitate its solution, Eq.~\eqref{nuclear_min_E} is recast as an unconstrained optimization problem by adding the standard Lagrangian term plus a quadratic penalty term \cite{bertsekas2014constrained} to yield the objective function
\begin{equation}\label{nuclear_min_lagrange}
L(\asb{Y}^{(i)},\asb{E},\boldsymbol{\Lambda},\mu) = \|\asb{Y}^{(i)}\|_* + \langle \boldsymbol{\Lambda}, \asb{M}^{(i)} - \asb{Y}^{(i)} - \asb{E} \rangle + \frac{\mu}{2} \|\asb{M}^{(i)} - \asb{Y}^{(i)} - \asb{E}\|_F
\end{equation}
where \(\|\cdot||_F\) denotes the Frobenius norm, \(\boldsymbol{\Lambda}\) are the Lagrange multipliers and \(\mu\) is a positive scalar. 
The ALM procedure is presented in Algorithm \ref{ALM_alg}, wherein Eq.~\eqref{nuclear_min_lagrange} is solved independently at each time instant. The algorithm solves a sequence of unconstrained subproblems, where the solution of the previous problem is used as the initial guess for the next one. First, the augmented Lagrangian is optimized with respect to \(\asb{Y}^{(i)}\) and \(\asb{E}^{(i)}\) via a  Proximal Gradient scheme \cite{beck2009fast} while keeping the the Lagrange multipliers \(\boldsymbol{\Lambda}\) fixed. The update on the singular values of \(\asb{Y}^{(i)}\) is performed using the soft-thresholding (shrinkage) operator 
\begin{equation}
    \mathcal{S}_\varepsilon[x] 
    \begin{cases}
    x - \varepsilon, & \text{if } x > \varepsilon, \\
    x + \varepsilon, & \text{if } x < -\varepsilon, \\
    0, & \text{otherwise},
    \end{cases}    
\end{equation}
and by enforcing \(P_{\bar{\Omega}}(\asb{E})=0\) such that \(P_{\bar{\Omega}}(\asb{Y}^{(i)})=0\). Next, \(\boldsymbol{\Lambda}\) is updated based on the degree of violation of the constraints  and the algorithm is repeated until convergence. The reader is directed to \cite{lin2010augmented} for more details about the derivation and the choice of the hyperparameters; see also \cite{hestenes1969multiplier, rockafellar1973multiplier, powell1978fast}.
\begin{algorithm}
    \caption{Augmented Lagrange Multipliers (ALM) Method based on \cite{lin2010augmented}}
    \label{ALM_alg}
    \begin{algorithmic}
        \State \textbf{Input:} observation set \( \Omega \), sampled entries \( \mathcal{P}_{\Omega}(\asb{M}^{(i)}) \)
        \State \textbf{Output:} \( \asb{Y}^{(i)}_k, \asb{E}_k \)
        \State \( \boldsymbol{\Lambda}_0 = 0; \quad \boldsymbol{E}_0 = 0; \mu_0>0;\rho>1; k=0\)
        \While{not converged}
            \State // solve: \( \asb{Y}^{(i)}_{k+1} = \arg \min_{\asb{Y}^{(i)}} L(\asb{Y}^{(i)}, \boldsymbol{E}_k, \boldsymbol{\Lambda}_k, \mu_k) \)
            \State \( [\boldsymbol{U}, \boldsymbol{S}, \boldsymbol{V}] = \mathrm{svd}(\asb{M}^{(i)} - \boldsymbol{E}_k + \mu_k^{-1} \boldsymbol{\Lambda}_k) \)
            \State \( \asb{Y}^{(i)}_{k+1} = \boldsymbol{U} \mathcal{S}_{\mu_k^{-1}}[\boldsymbol{S}] \boldsymbol{V}^T \) \Comment{\( \mathcal{S} \): soft thresholding (shrinkage) operator}
            \State // solve: \( \boldsymbol{E}_{k+1} = \arg \min_{\pi_{\Omega}(\boldsymbol{E})=0} L(\asb{Y}^{(i)}_{k+1}, \boldsymbol{E}, \boldsymbol{\Lambda}_k, \mu_k) \)
            \State \( \boldsymbol{E}_{k+1} = P_{\bar{\Omega}}(\asb{M}^{(i)} - \asb{Y}^{(i)}_{k+1} + \mu_k^{-1} \boldsymbol{\Lambda}_k) \)
            \State \( \boldsymbol{\Lambda}_{k+1} = \boldsymbol{\Lambda}_k + \mu_k (\asb{M}^{(i)}_j - \asb{Y}^{(i)}_{k+1} - \boldsymbol{E}_{k+1}); \quad \mu_{k+1} = \rho \mu_k \)
            \State \( k \gets k + 1 \)
        \EndWhile
    \end{algorithmic}
\end{algorithm}

Note that matrix completion algorithms
have been used recently in various civil engineering applications. For instance, a structural response denoising strategy was proposed in \cite{yang2014blind} by decomposing the data matrix into a low-rank matrix and a sparse outlier representation. In \cite{yang2015data} the same authors employed 
a low-rank matrix scheme to identify large-scale structural seismic and typhoon responses. The reader is referred to the work in \cite{yang2016harnessing} and the references therein for a broader discussion. 

Overall, the formulation of Eq.~\eqref{nuclear_min} leads to a convex optimization problem, which can be solved efficiently and can account for a considerable degree of missing data. Nevertheless, the methodology has several limitations. First, the low rank assumption may not hold in practice when the data are collected over a large spatial domain with significant spatial variability and/or when the measurement separation distance is large compared to the wind field spatial coherence function. Second, this approach accounts for spatial correlations at each time instant independently, neglecting the temporal correlations of the wind field. Third, the method typically assumes uniform random sampling (e.g., Bernoulli sampling pattern) of the data, which may not align with sensor deployment scenarios in engineering applications. Fourth, it does not provide an explicit expansion basis and, therefore, lacks interpretability. Finally, the method does not quantify the uncertainty associated with the estimates, which limits its applicability in an engineering context.

\subsection{Nonparametric Bayesian Dictionary Learning}\label{DL_section}

Ideally, a method that ``learns" a sparse expansion directly from the data would be desirable. The related field of dictionary learning (DL) offers several algorithms for finding a linearly independent set of vectors \cite{engan1999method} together with the associated sparse coefficients. For uncertainty quantification, DL can also be formulated from a Bayesian perspective \cite{olshausen1997sparse,aharon2006k}. In its general form, the DL problem is given as
\begin{equation}
    \min_{A, W} \; \| \mathbf{Z} - \mathbf{A} \mathbf{W} \|_F^2 \quad \text{subject to} \quad \| \asb{W}_i \|_0 \leq k_0 \; \forall \; i = 1, \dots, L
\end{equation}
where \(\mathbf{Z}\) is a collection of \(L\) time histories such that \(\mathbf{Z} = [\mathbf{x}_1, \mathbf{x}_2, \dots, \mathbf{x}_L] \in \mathbb{R}^{MN \times L}\). Note that the majority of DL methods require a collection of $L$ complete data snapshots, which is typically not feasible in wind engineering applications due to sensor deployment constraints. Further, this family of methods typically assume that the size of the dictionary is determined \textit{a priori} and that the noise/residual variance is known; see, for instance, the work in \cite{aharon2006k} where \(||\mathbf{Z} - \mathbf{A} \mathbf{W} \|_F^2<\epsilon\). 

To address these limitations, recent results in unsupervised learning and non-parametric Bayesian statistics \cite{hjort2010bayesian} are leveraged herein to obtain a sparse wind field data representation from limited observations. This family of methods models the observations as a linear combination of (potentially) infinite features whose number is not known \textit{a priori}. Within this family, two general methods have been proposed. The first is the Dirichlet process mixture model, which assumes a set of mutually exclusive clusters, with each datapoint assigned to a latent class \cite{ferguson1973bayesian}. The second method is the beta-Bernoulli formulation, which is more general, and models the process using a combination of latent features where each feature is a Bernoulli random variable with parameters sampled from a beta distribution \cite{paisley2009nonparametric, broderick2012beta}. This model was recently used to cast factor analysis as a dictionary learning problem to denoise and interpolate incomplete images. Specifically, the methodology proposed in \cite{paisley2009nonparametric,zhou2011nonparametric}, also known as Beta Process Factor Analysis (BPFA), employs a hierarchical Bayesian model to represent image patches using low-dimensional dictionary elements. In this approach, priors are used to model the dictionary, the coefficients, and the noise, which allows one to estimate noise statistics and to adopt flexible priors based on the application.

In this work, the BPFA algorithm proposed in \cite{zhou2011nonparametric} is adapted for joint space-time wind field data extrapolation. Consider the original data \( \asb{Y} \in \mathbb{R}^{M_x \times M_z \times N}\), which are partitioned into a set of vectorized overlapping voxels \(\{\mathbf{y}_i\}_{i=1}^{n_{total}}\) such that \(\mathbf{y}_i \in \mathbb{R}^{m_x m_z N}\) where \(m_x\) and \(m_z\) denote the size of the block in each dimension. The total number of blocks is given as \(n_{total} = (\frac{M_x - m_x}{\Delta_x}+1)(\frac{M_z - m_z}{\Delta_y}+1)\) where \(\Delta_x\) and \(\Delta_z\) are the strides in each dimension, respectively. Next, each data block is modeled as
\begin{equation}
    \mathbf{y}_i = \mathbf{\Sigma}_i(\mathbf{Dw}_i+\boldsymbol{\epsilon}_i), \mathbf{D} \in \mathbb{R}^{m_x m_z N\times K}, \text{ with } K \rightarrow \infty
    \label{CS_BPFA}
\end{equation}
in which \(\mathbf{\Sigma}_i \in R^{{m_x m_z N} \times m_x m_z N}\) is the observation matrix and \(\boldsymbol{\epsilon}_i \in \mathbb{R}^{m_x m_z N}\) is the measurement error vector. Additionally, prior distributions are placed over each model variable. Specifically, each atom \(\{\mathbf{d}_k\}_{k=1}^K\), corresponding to the columns of \(\mathbf{D}\), is sampled by a Gaussian distribution in the form
\begin{align*}
    \mathbf{d}_k &\sim \mathcal{N}(0, P^{-1} \mathbf{I}_P)
\end{align*}
where \(\mathbf{I}_P\) is a \(P \times P\) identity matrix and \(P=m_x m_z N\) is the number of rows for each atom. The coefficients \(\mathbf{w}_i\) are decomposed as
\begin{align*}
    \mathbf{w}_i &= \mathbf{z}_i \pdot \mathbf{s}_i
\end{align*}
where \(\pdot\) denotes the Hadamard product, \(\mathbf{s}_i \in R^K\) denotes the magnitude of the coefficients, and \(\mathbf{z}_i \in \{0, 1\}^K\) is a binary vector where each element can take the value \(0\) or \(1\) to indicate the ``active set" of columns in \(\mathbf{D}\) and ensures sparsity. 
The vector \(\mathbf{z}_i\) is generated by a Bernoulli process parameterized by a two-parameter Beta distribution in the form
\begin{equation}\label{BP_prior}
    \mathbf{z}_i \sim \prod_{k=1}^K \text{Bernoulli}(\pi_k), \quad \boldsymbol{\pi} \sim \prod_{k=1}^K \text{Beta}\left(\frac{a}{K}, \frac{b(K-1)}{K}\right)
\end{equation}
in which the \(\mathbf{z}_i\) and \(\boldsymbol{\pi}\) are drawn independently from their respective distributions and \(a,b\) are parameters of the Beta distribution. This sampling scheme leads to an increasingly sparse solution as \(i\) increases, such that each subsequent block adds a diminishing amount of new information. This promotes sparsity in an analogous manner to \(\ell_0\) minimization, where the majority of the coefficients are forced to be exactly zero. Further, dictionary atoms that are increasingly likely to be nonzero at each iteration are consistently used in the signal representation. More information on the Beta process construction can be found in \cite{zhou2011nonparametric,griffiths2011indian,paisley2009nonparametric}. The coefficients \(\mathbf{s}_i\) are sampled from a Gaussian distribution with zero mean and precision \(\gamma_s\) in the form
\begin{align}\label{s_prior}
    \mathbf{s}_i & \sim \mathcal{N}(0, \gamma_s^{-1} \mathbf{I}_K) \\
    \gamma_s  & \sim \text{Gamma}(c,d)
\end{align}
The error is also modeled as a Gaussian random variable parameterized by a Gamma-distributed variance as follows
\begin{align}\label{epsilon_prior}
    \mathbf{\epsilon}_i &\sim \mathcal{N}(0, \gamma_{\epsilon}^{-1} \mathbf{I}_P) \\
    \gamma_{\epsilon} &\sim \text{Gamma}(e, f)
\end{align}
where \(c, d,e,f\) are distribution parameters. This allows for estimation of the noise variance as a measure of uncertainty in the estimates. Overall, the hierarchical Bayesian model leads to the following joint probability distribution function (PDF)
\begin{align}\label{BP_posterior}
    &P(\mathbf{\asb{Y}}, \boldsymbol{\Sigma}, \mathbf{D}, \mathbf{Z}, \mathbf{S}, \boldsymbol{\pi}, \gamma_s, \gamma_\epsilon) \notag \\
    &= \prod_{i=1}^N \mathcal{N}( \mathbf{y_i} ; \boldsymbol{\Sigma}_i \mathbf{D} (\mathbf{s}_i \odot \mathbf{z}_i), \gamma_\epsilon^{-1} \mathbf{I}_{\|\Sigma_i\|_0}) \mathcal{N}(s_i; 0, \gamma_s^{-1} \mathbf{I}_K) \times \notag \\
    &\quad \prod_{k=1}^K \mathcal{N}(\mathbf{d_k}; 0, P^{-1} \mathbf{I}_P) \text{Beta}(\pi_k; a, b) \times \notag \\
    &\quad \prod_{i=1}^N \prod_{k=1}^K \text{Bernoulli}(z_{ik}; \pi_k) \times\notag \\
    &\quad \Gamma(\gamma_s; c, d) \Gamma(\gamma_\epsilon; e, f)
\end{align}
The full posterior can be inferred using Gibbs sampling where each variable is approximated iteratively by sampling from the conditional distribution given by the remaining variables. Taking into account that each consecutive density function belongs to the conjugate-exponential family, the Gibbs update equations can be analytically derived for each hyperparameter. These steps can be found in \cite{zhou2011nonparametric} and have been included in \ref{appendix} for completeness.

\section{Wind field extrapolation in two spatial dimensions}\label{section2}
In this section, the low rank matrix recovery and the Bayesian dictionary learning framework are employed to extrapolate missing simulated wind velocity records in a two-dimensional domain. To assess its efficacy in obtaining statistically compatible time histories, the wind field is simulated as a homogeneous stochastic wave using the spectral representation method~\cite{benowitz2015simulation}. This facilitates the efficient simulation of realizations corresponding to a large number of points in the spatial domain, while circumventing the need for costly cross-PSD related calculations; see \cite{deodatis1996simulation} for details on simulation of multi-variate (vector) processes. Here, relevant statistical quantities can be computed analytically and compared to their extrapolated counterparts. The details of the Monte Carlo simulation (MCS) procedure are briefly detailed in the following section.

\subsection{Wind field simulation using the stochastic wave spectral representation method}\label{section21}
Following the derivation in \cite{chen2018simulation,benowitz2015simulation}, the PSD of the homogeneous wind field in the frequency-wavenumber domain can be expressed as 
\begin{equation}\label{WF_transform}
    \centering 
    S(\kappa_x, \kappa_z, \omega)=\frac{1}{(2 \pi)^2} \int_{-\infty}^{\infty} \int_{-\infty}^{\infty} S_0(\omega)\gamma(\xi_x, \xi_z, \omega) e^{-i \kappa_x \xi_xt} e^{-i \kappa_z \xi_z} \mathrm{d} \xi_x  \mathrm{d} \xi_z
\end{equation}
in which $S_0(\omega)$ is the auto-PSD, $\gamma(\cdot)$ is the spatial coherence function, and \(\xi_x, \xi_z \) denote the spatial separation distances \(\xi_x = x_i-x_j\), \(\xi_z = z_i-z_j \ \forall i = 1,\dots,M_x, \ \forall j = 1,\dots,M_z\) in the \(x\) and \(z\) directions, respectively. Consider the following coherence function 
\begin{equation}\label{LF_spec}
    \gamma(\xi_x, \xi_z, \omega)=\exp \left(-\frac{1}{2 \pi U_{10}}|\omega|\sqrt{C_{1x}^2\xi_x^2 - C_{1z}^2\xi_z^2}\right)
\end{equation}
where \(C_{1x}\) and \(C_{1z}\) are the exponential decay coefficients corresponding to the horizontal and vertical directions, respectively and \(U_{10}\) is the mean wind velocity at a height of \(10\)m.  
The Davenport auto-PSD model \cite{davenport1961spectrum} is adopted, which takes the form
\begin{eqnarray}\label{Davenport}
    \centering 
         &S_0(\omega) =  2.0 u_{*}^{2} \frac{\left(\frac{1200}{2 \pi U_{10}} \omega\right)^{2}}{|\omega|\left(1+\left(\frac{1200}{2 \pi U_{10}} \omega\right)^{2}\right)^{4 / 3}}
\end{eqnarray}
in which \(u_{*}\) denotes the shear flow velocity. Substituting Eqs.~\eqref{LF_spec} and \eqref{Davenport} into Eq.~\eqref{WF_transform} and calculating the double integral yields 
\begin{eqnarray}\label{WF_Spec_2D}
    \begin{aligned}
        S\left(\kappa_{x}, \kappa_{z}, \omega\right) 
        % S_0(\omega) \cdot \gamma\left(\kappa_{y}, \kappa_{z}, \omega\right)&\\ 
        = &\frac{u_{*}^{2}}{\pi C_{1 x} C_{1 z}\left(\frac{1}{2 \pi U_{10}}|\omega|\right)^{2}} \frac{\left(\frac{1200}{2 \pi U_{10}} \omega\right)^{2}}{|\omega|\left(1+\left(\frac{1200}{2 \pi U_{10}} \omega\right)^{2}\right)^{4 / 3}}\\
         &\times \frac{1}{\left(1+\left[\left(\frac{1}{C_{1 x}} \kappa_{x}\right)^{2}+\left(\frac{1}{C_{1 z}} \kappa_{z}\right)^{2}\right] /\left(\frac{1}{2 \pi U_{10}}|\omega|\right)^{2}\right)^{\frac{3}{2}}}
    \end{aligned}        
\end{eqnarray}
Next, sample realizations of the velocity field are generated using the Spectral Representation Method (e.g., \cite{benowitz2015simulation}, \cite{chen2018simulation}) as
\begin{eqnarray}\label{SRM_2D}
    \begin{aligned}
        \mathrm{v}(x, z, t)=& \sum_{i=1}^{N_{\kappa_{x}}} \sum_{j=1}^{N_{\kappa_{z}}} \sum_{k=1}^{N_{\omega}} \sqrt{4 S\left(\kappa_{i}^{(x)}, \kappa_{j}^{(z)}, \omega_{k}\right) \Delta \kappa^{(x)}_i \Delta \kappa^{(z)}_{j} \Delta \omega^{(k)}} \\
        & \cdot\left[\cos \left(\kappa_{i}^{(x)} x+\kappa_{j}^{(z)} z+\omega_{k} t+\varphi_{i j k}^{(1)}\right)\right.\\
        &+\cos \left(\kappa_{i}^{(x)} x+\kappa_{j}^{(z)} z-\omega_{k} t+\varphi_{i j k}^{(2)}\right) \\
        &+\cos \left(\kappa_{i}^{(x)} x-\kappa_{j}^{(z)} z+\omega_{k} t+\varphi_{i j k}^{(3)}\right) \\
        &\left.+\cos \left(\kappa_{i}^{(x)} x-\kappa_{j}^{(z)} z-\omega_{k} t+\varphi_{i j k}^{(4)}\right)\right]
        \end{aligned}
\end{eqnarray}
where \(\kappa_{i}^{(x)}=i \Delta \kappa_{x}, i=1,2, \ldots, N_{\kappa_{x}}\) and \(\kappa_{j}^{(z)}=j \Delta \kappa_{z}, j=1,2, \ldots, N_{\kappa_{z}}\) are the discretized wavenumbers in the \(x\) and \(z\) directions, respectively, \(\omega_k=k\Delta\omega, \quad k=1, 2, \dots, N_{\omega}\) and \(\varphi_{i j k}^{(1)}\), \(\varphi_{i j k}^{(2)}\), \(\varphi_{i j k}^{(3)}\) and \(\varphi_{i j k}^{(4)}\) represent four different sets of independent random phase angles uniformly distributed in \([0,2 \pi]\).  It is noted that the Davenport PSD of Eq.~\eqref{Davenport} exhibits a singularity at the origin. This is addressed using a frequency shift scheme described in \cite{zerva1992seismic,benowitz2015simulation}. Adopting an uneven discretization scheme, with sampling density proportional to the magnitude of the PSD, such as the one found in ~\cite{chen2001atomic}, can also reduce significantly the computational effort.

Fifty two-dimensional wind velocity field realizations are generated with the following simulation parameters: total time \(T_0 = M\Delta t = 511.62\) sec.; upper cut-off frequency \(w_u = N_\omega \Delta\omega = 8\pi\) rad/s; time increment \(\Delta t={\pi}/{w_u} = 0.0125\) sec.; frequency increment \(\Delta\omega = {2\pi}/{T_0} = 0.0123\) \(1/s\); upper cut-off wavenumbers \(\kappa_{u_x} = \kappa_{u_z} = \pi\) rad/m; wavenumber increments  \(\Delta \kappa_x = \Delta \kappa_z = 0.002\) rad/m; exponential decay coefficients \(C_{1x} = C_{1z} = 7\); mean wind velocity at \(10m\) \(U_{10} = 31.88\) m/s; shear flow velocity \(u_{*}=1.691\) m/s. Each wind field comprises time histories simulated at \(100\) points on a square grid with a separation distance of \(10 \) m between points in each direction as illustrated in Figure~\ref{Schematic_2D}. 

\subsection{Velocity prediction with dictionary learning}\label{numerical_example_1}
 
Sixty of the \(100\) time histories from each field are randomly removed from the data set and considered to have unknown velocity time histories. Extrapolating to the unknown points is cast as a time-dependent matrix completion problem by applying the ALM and Bayesian dictionary learning approaches described above. 
% in Fig.~\ref{ALM_alg}. 
Despite the high degree of missing data (\(60\%\)), minimizing the nuclear norm is expected to yield reasonable results considering the correlated structure of the points in the \(2D\) spatial domain given by Eq.~\eqref{WF_Spec_2D}. 
For DL, the records are partitioned in the spatial domain by considering consecutive overlapping blocks with block sizes \(m_x = m_z = 4\), stride \(\Delta_x = \Delta_y = 1\) and dictionary size of \(K=512\). Each block is vectorized to account for the complete time history, which yields a significant advantage compared to the matrix completion approach which ignores time dependence. The following values are used to initialize the hyperparameters of the distributions in Eqs.~\eqref{BP_prior}~-~\eqref{epsilon_prior}: \(\,a_0 = 1\,\) \(\,b_0 = 1\,\) \(\,c_0 = 10^{-6}\,\) \(\,d_0 = 10^{-6}\,\) \(\,e_0 = 10^{-6}\,\) \(\,f_0 = 10^{-6}\).

\begin{figure}[!ht]
    \includegraphics[width=\textwidth]{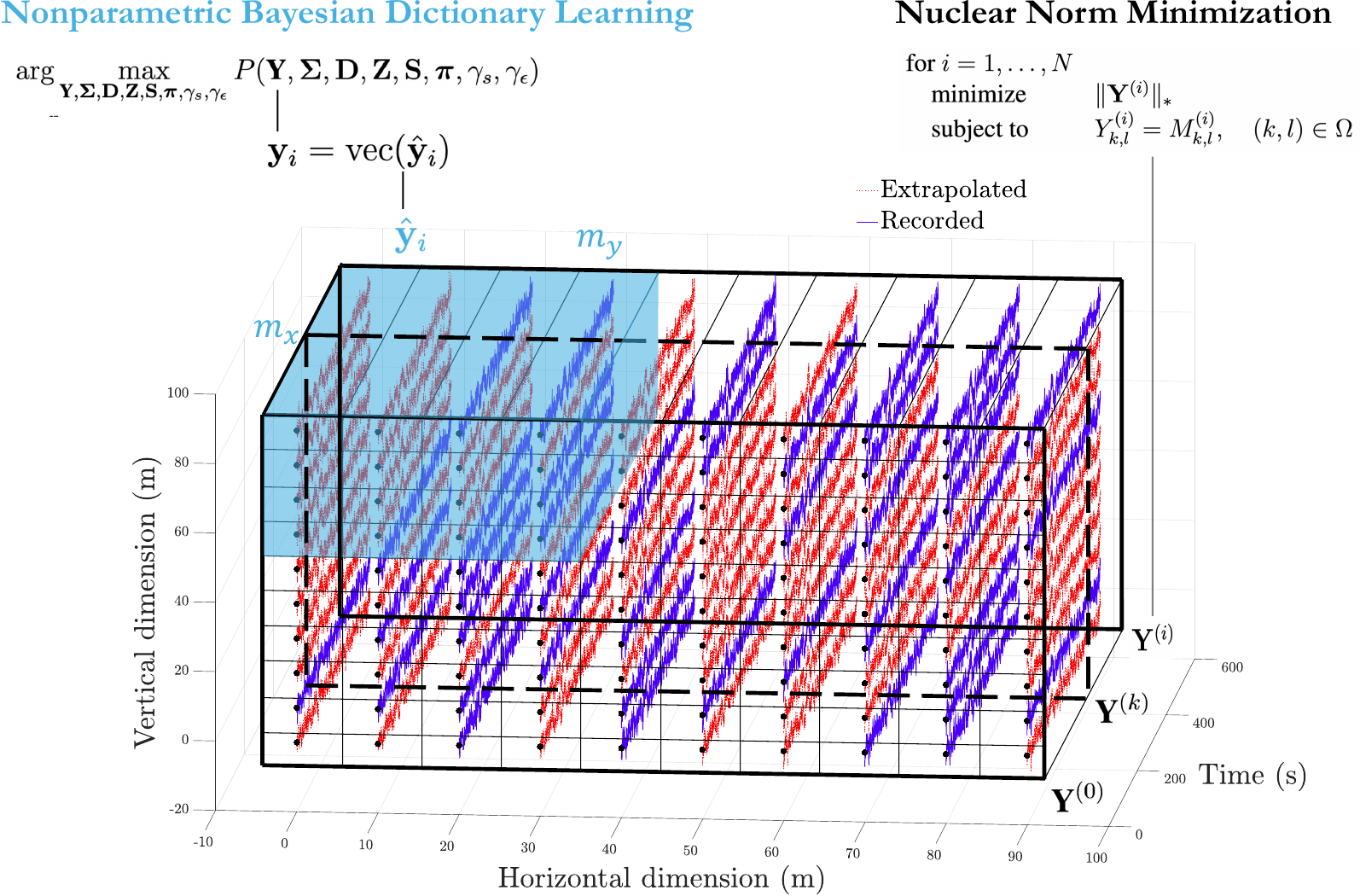}
    \caption{Schematic representation of the extrapolation scheme in two dimensions via nuclear norm minimization and Bayesian dictionary learning; extrapolating to 60 grid points based on 40 measurement locations}
    \label{Schematic_2D}
\end{figure}

The extrapolation scheme for both optimization procedures is shown in Fig.~\ref{Schematic_2D}.
The missing records are extrapolated for \(N\) independent realizations using both sparse representation methods. The posterior distribution in Eq.~\eqref{BP_posterior} is then sampled to estimate uncertainties in the predictions. Finally, stochastic field statistics, i.e., the PSD, the cross-correlation and the coherence function are estimated based on the ensemble average of the extrapolated time-histories. 

\begin{figure}[!ht]
    \centering
    \includegraphics[width=\textwidth]{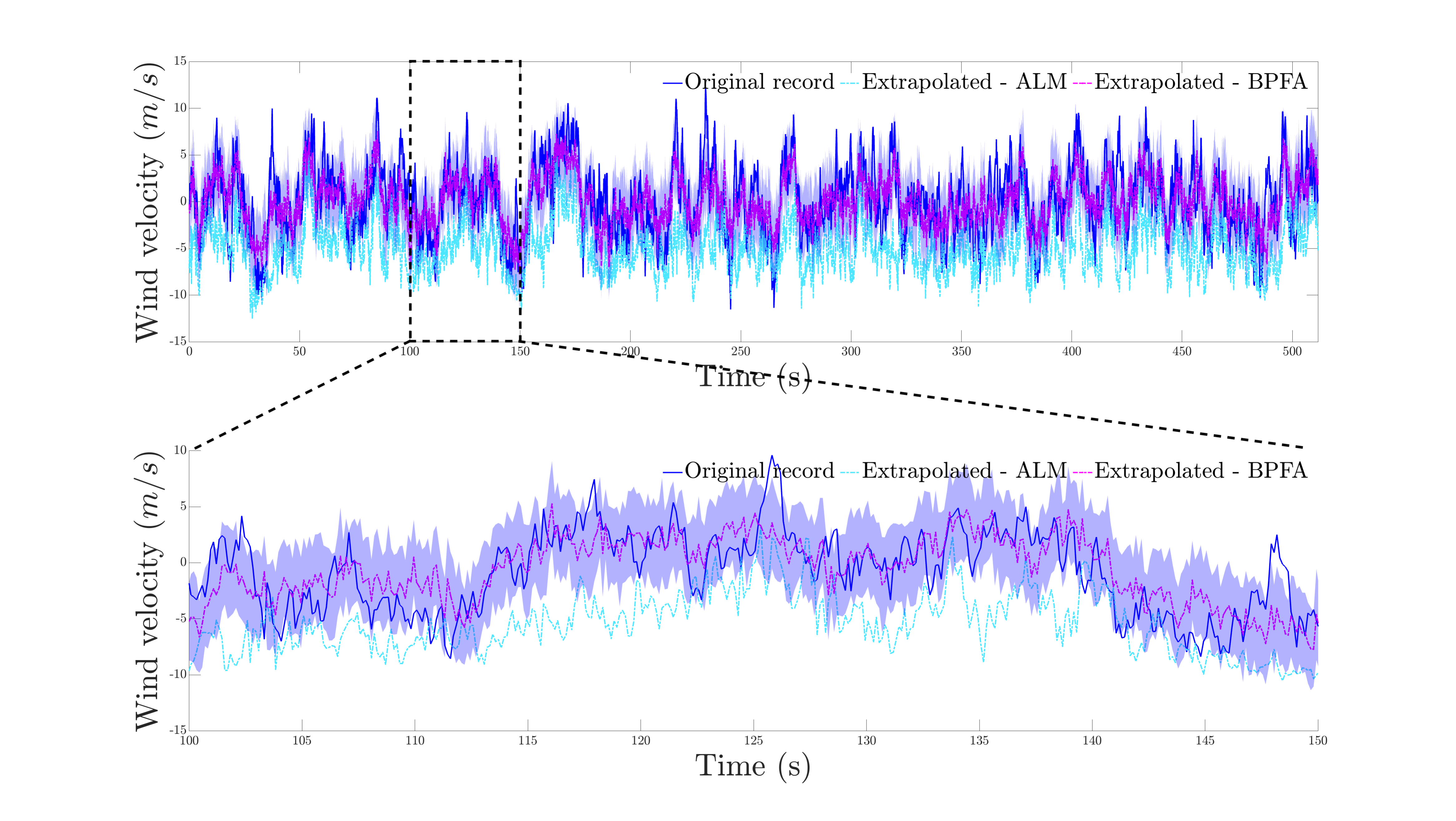}
    \caption{Extrapolated time history at point \((0,70)\) from Fig.~\ref{Schematic_2D} using ALM and BPFA. Top figure: Full record. Bottom figure: 50 second segment of the record showing detailed fluctuations. }\label{sample_0_70}
\end{figure}

\begin{figure}[!ht]
    \centering  
    \includegraphics[width=\textwidth]{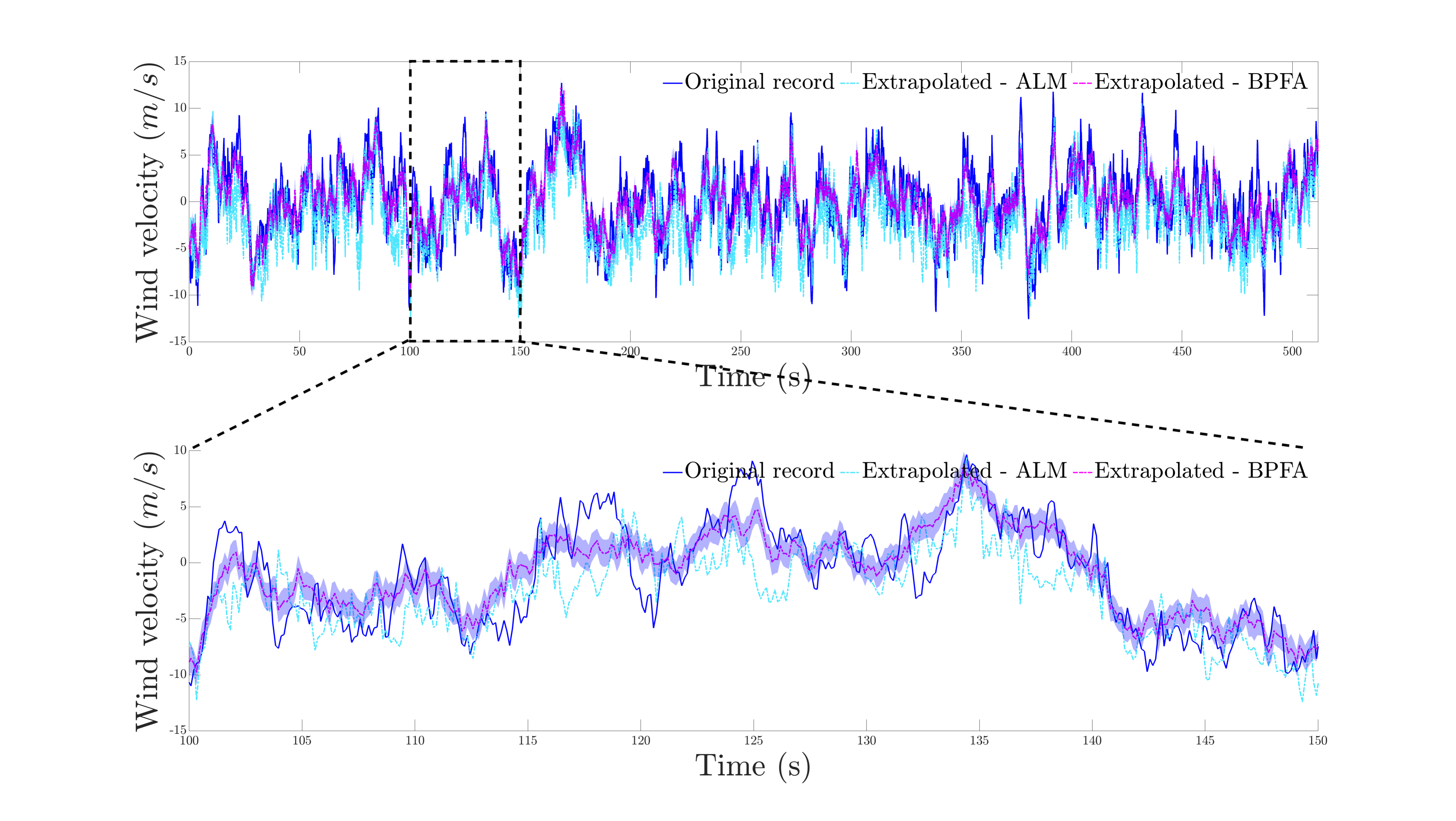}
    \caption{Extrapolated time history at point \((50,30)\) from Fig.~\ref{Schematic_2D} using ALM and BPFA.  Top figure: Full record. Bottom figure: 50 second segment of the record showing detailed fluctuations.}\label{sample_50_30}
\end{figure}

\begin{figure}[!ht]
    \centering  
    \includegraphics[width=0.7\textwidth]{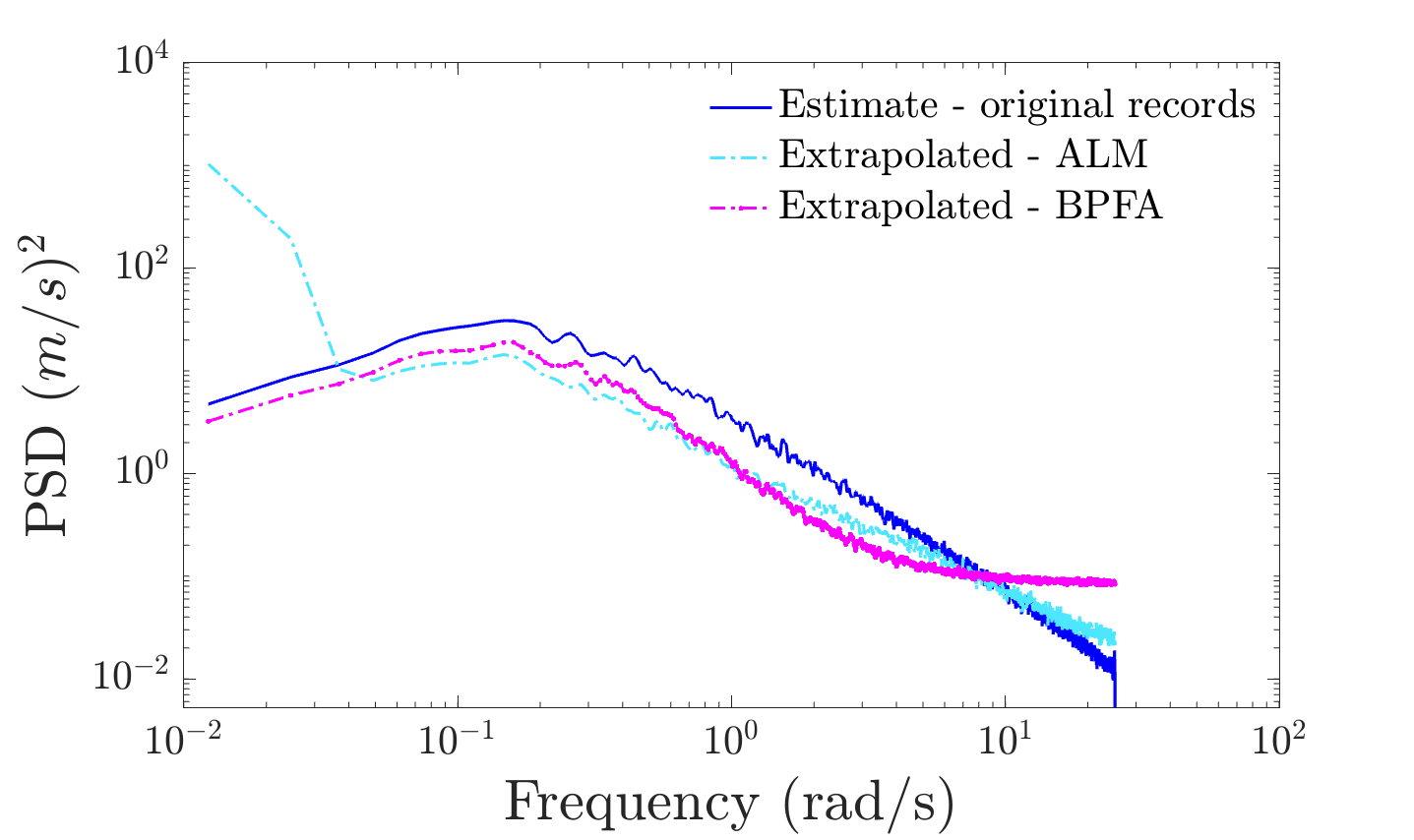}
    \caption{Comparison of the estimated PSD at point \((0,70)\) from Fig.~\ref{Schematic_2D} based on the ensemble average using the ALM algorithm and the BPFA approach.}
    \label{PSD_0_70_log}
\end{figure}

\begin{figure}[!ht]
    \centering  
    \includegraphics[width=0.7\textwidth]{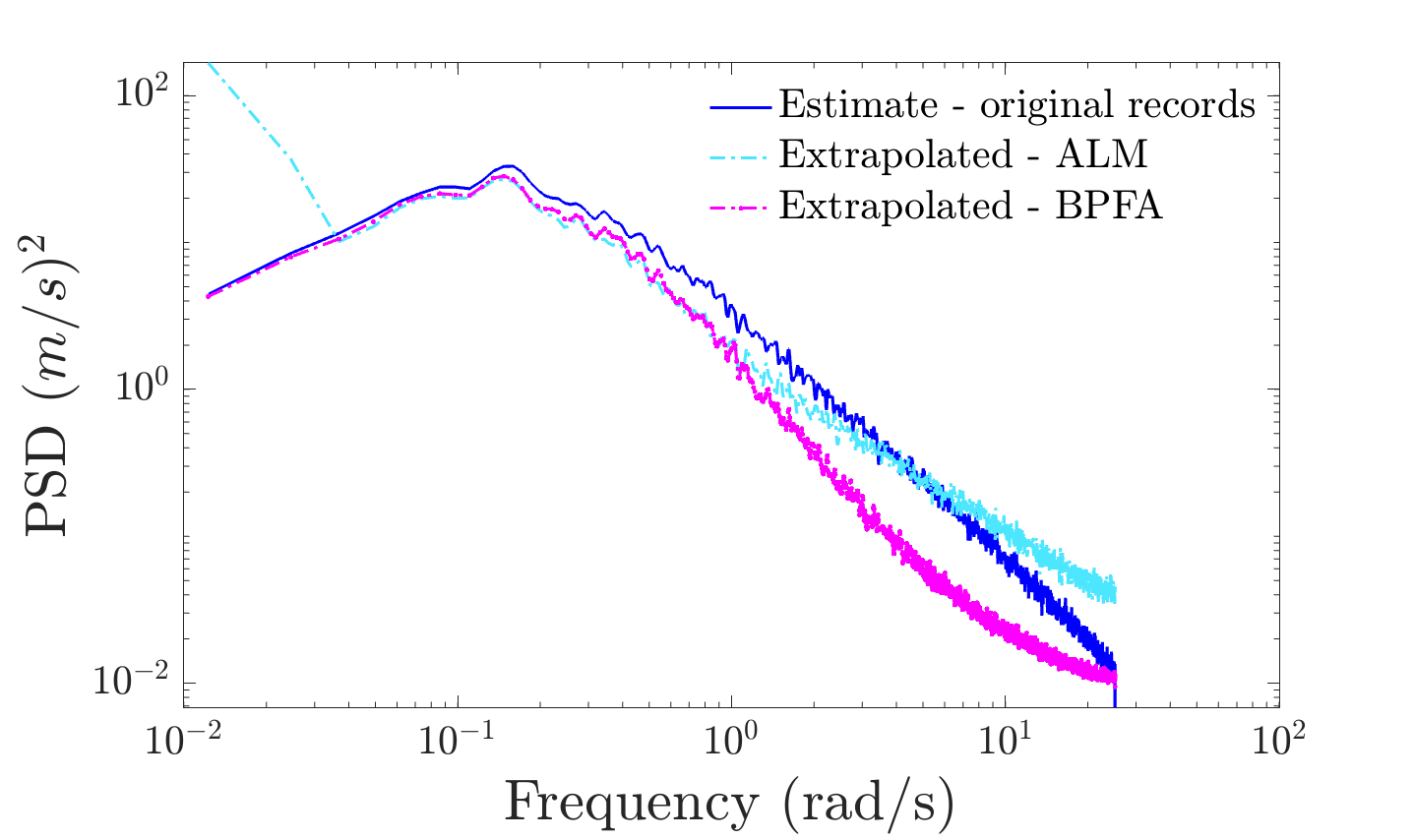}
    \caption{Comparison of the estimated PSD at point \((50,30)\) from Fig.~\ref{Schematic_2D} based on the ensemble average using the ALM algorithm and the BPFA approach.}\label{PSD_50_30_log}
\end{figure}

\begin{figure}
    \centering
    \includegraphics[width=0.7\textwidth]{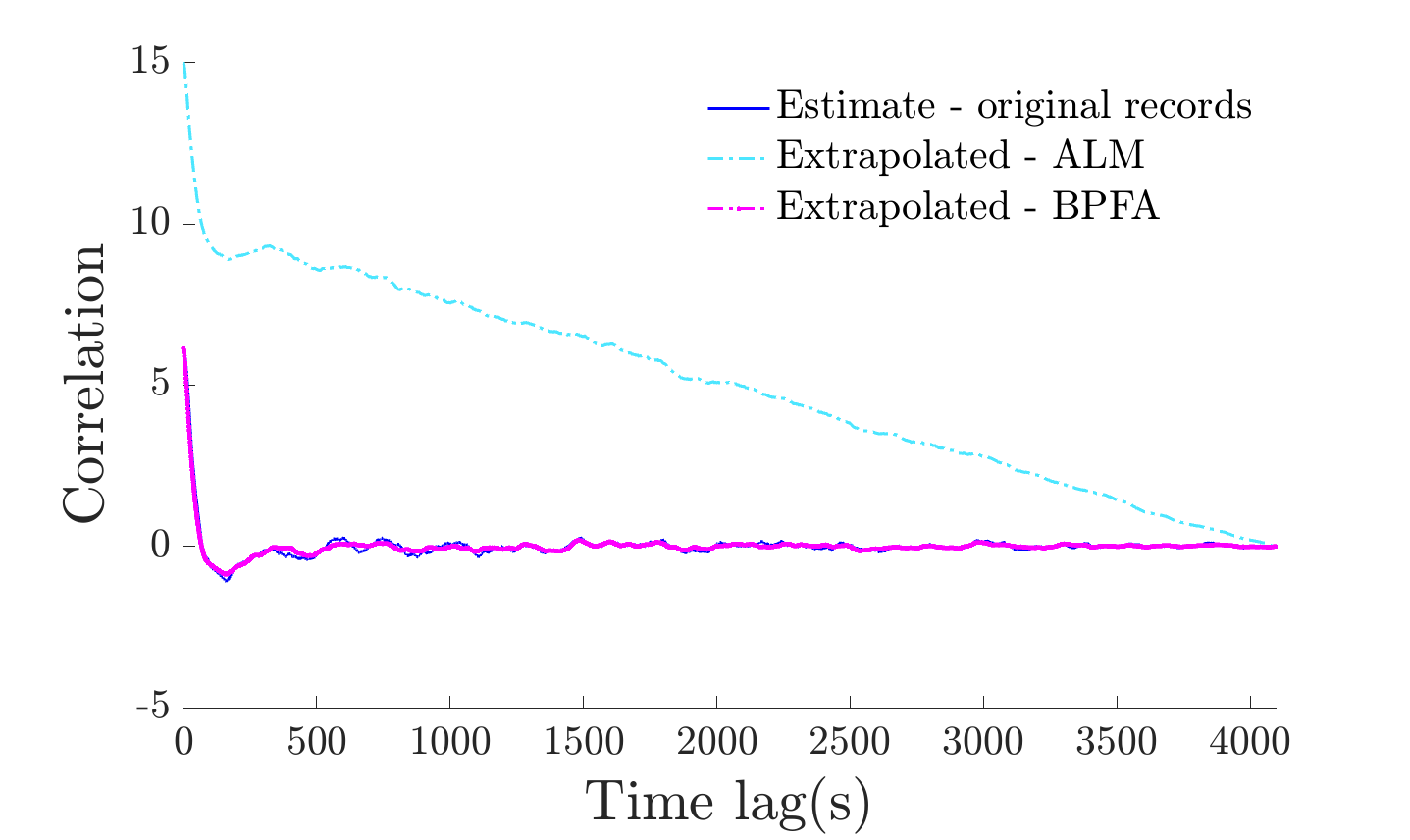}
    \caption{Comparison of the estimated cross-correlation between points \((0,70)\) and \((50,30)\) from Fig.~\ref{Schematic_2D} based on the ensemble average using the ALM algorithm and the BPFA approach.}\label{Corr_comparison_(0_70_50_30)}
    \label{Corr_comparison_(0_70)}
\end{figure}

\begin{figure}[!ht]
    \centering
    \includegraphics[width=0.7\textwidth]{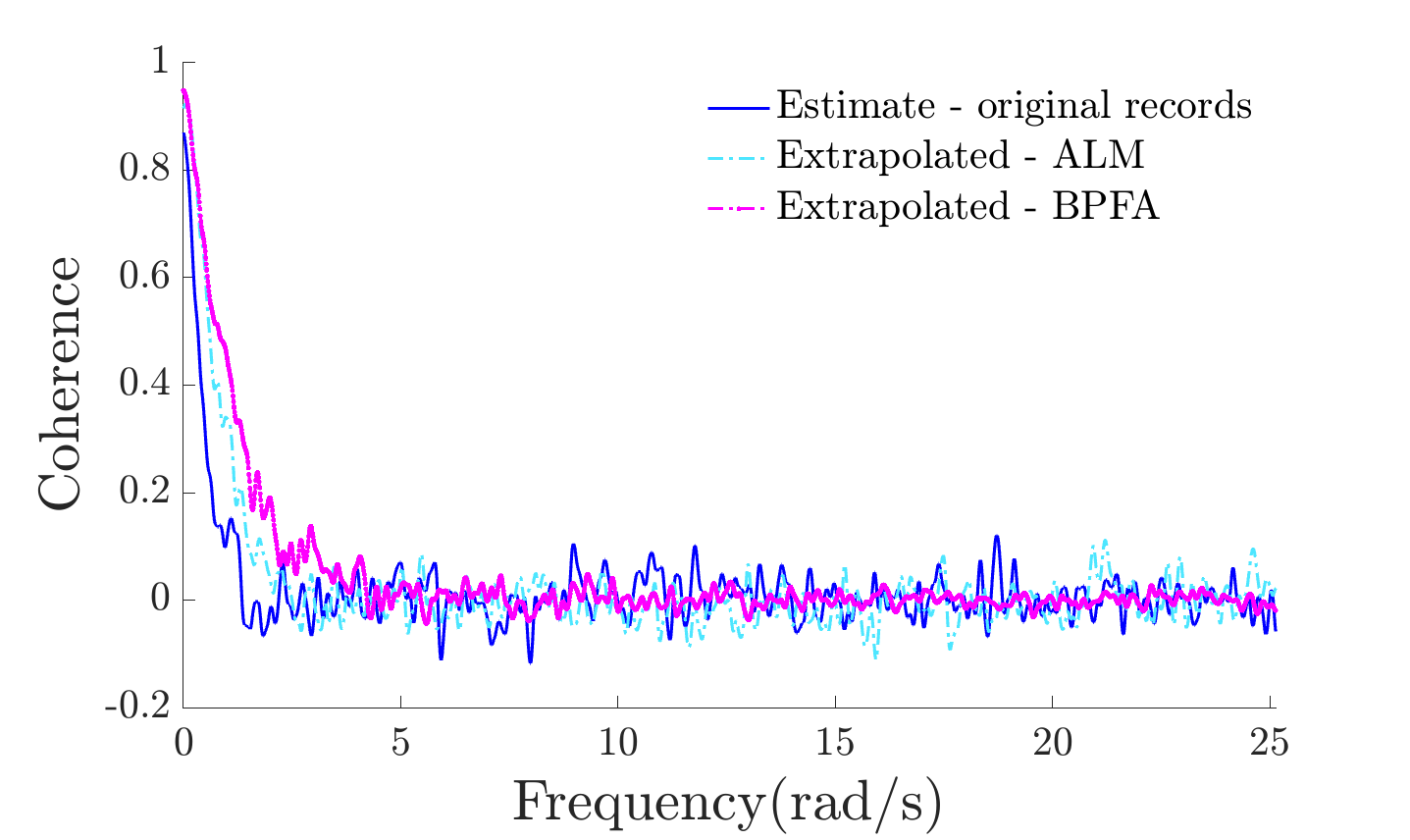}
    \caption{Comparison of the estimated cross-coherence between points \((0,70)\) and \((50,30)\) from Fig.~\ref{Schematic_2D} based on the ensemble average using the ALM algorithm and the BPFA approach.}\label{Coh_comparison_(0_70_50_30)}
\end{figure}

\begin{figure}[!ht]
    \centering
    \includegraphics[width=0.6\textwidth]{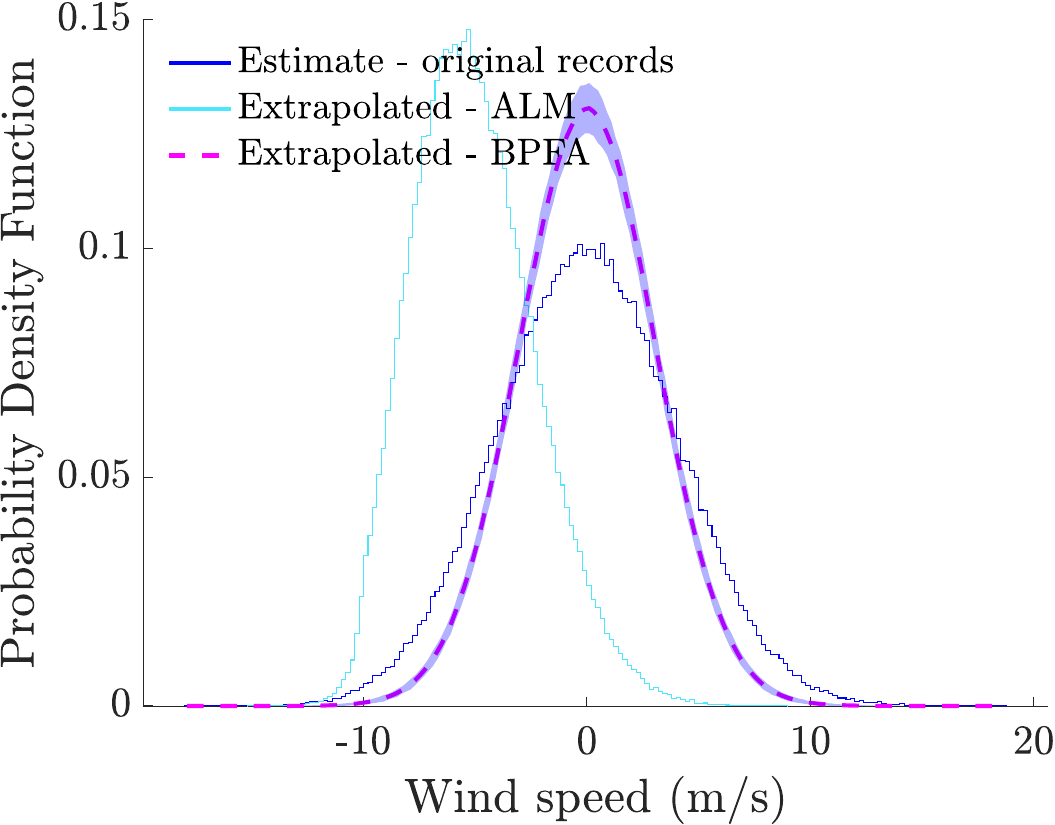}
    \caption{Estimated PDF of the extrapolated time histories at point \((0,70)\) from Fig.~\ref{Schematic_2D} based on the ensemble of simulations using the ALM algorithm and the BPFA  approach.}\label{Hist_comparison_(0_70)}
\end{figure}

\begin{figure}[!ht]
    \centering\includegraphics[width=0.6\textwidth]{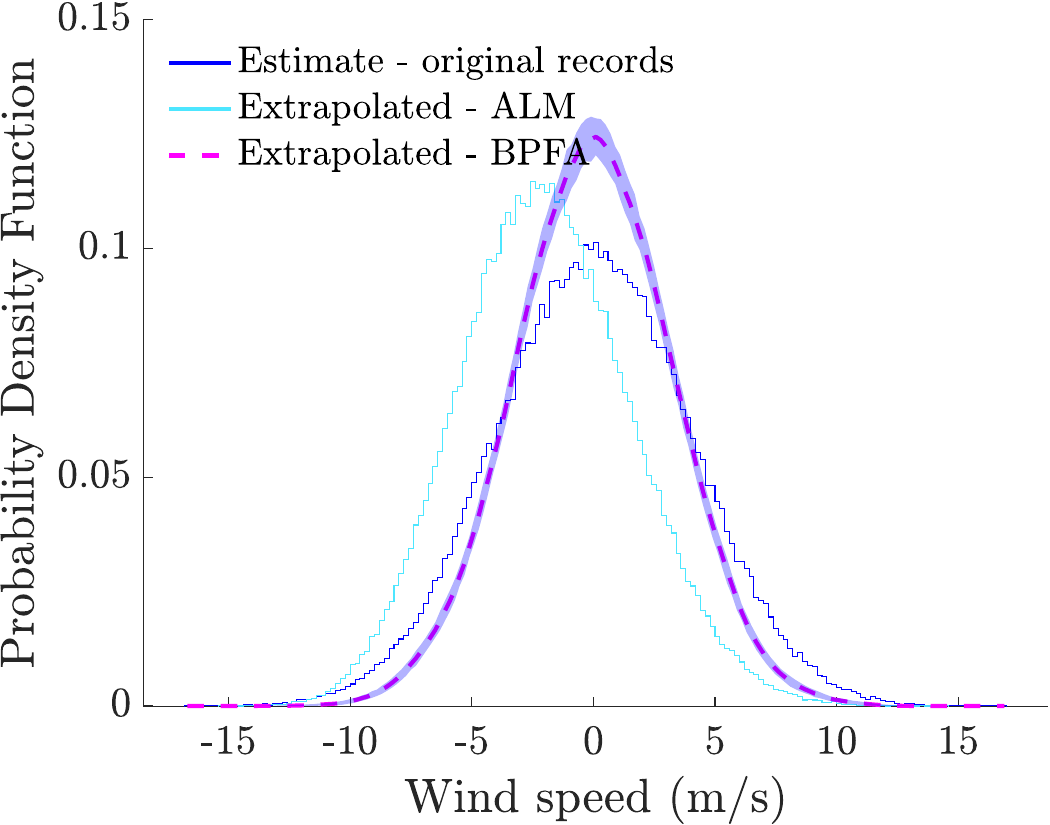}
    \caption{Estimated PDF of the extrapolated time histories at point \((50,30)\) from Fig.~\ref{Schematic_2D} based on the ensemble of simulations using the ALM algorithm and the BPFA approach.}\label{Hist_comparison_(50_30)}
\end{figure}

Two representative extrapolated records, associated with points \((0,70)\) and \((50,30)\), are shown in Fig.~\ref{sample_0_70} and Fig.~\ref{sample_50_30}, respectively. The BPFA approach demonstrates considerable performance in reconstructing the time histories, particularly in capturing the temporal features in greater detail. The estimated PSDs are shown in Fig.~\ref{PSD_0_70_log}~--~\ref{PSD_50_30_log}, cross-correlations are shown in Fig.~\ref{Corr_comparison_(0_70_50_30)}, and the coherence functions are shown in Fig.~\ref{Coh_comparison_(0_70_50_30)}. It is seen that the BPFA method outperforms the ALM in estimating the PSD peaks, whereas the discrepancies are observed primarily in low-amplitude, high frequency content. Although both methods yield comparable estimates of the coherence function, the BPFA algorithm yields a better estimate of the correlation between the two missing spatial locations.  Moreover, the estimated probability density functions (PDFs) of the time histories are shown in Fig.~\ref{Hist_comparison_(0_70)} and Fig.~\ref{Hist_comparison_(50_30)} where the BPFA captures the mode of the PDF more accurately compared to the ALM.



\begin{figure}[!ht]
    \centering
    \includegraphics[width=0.7\textwidth]{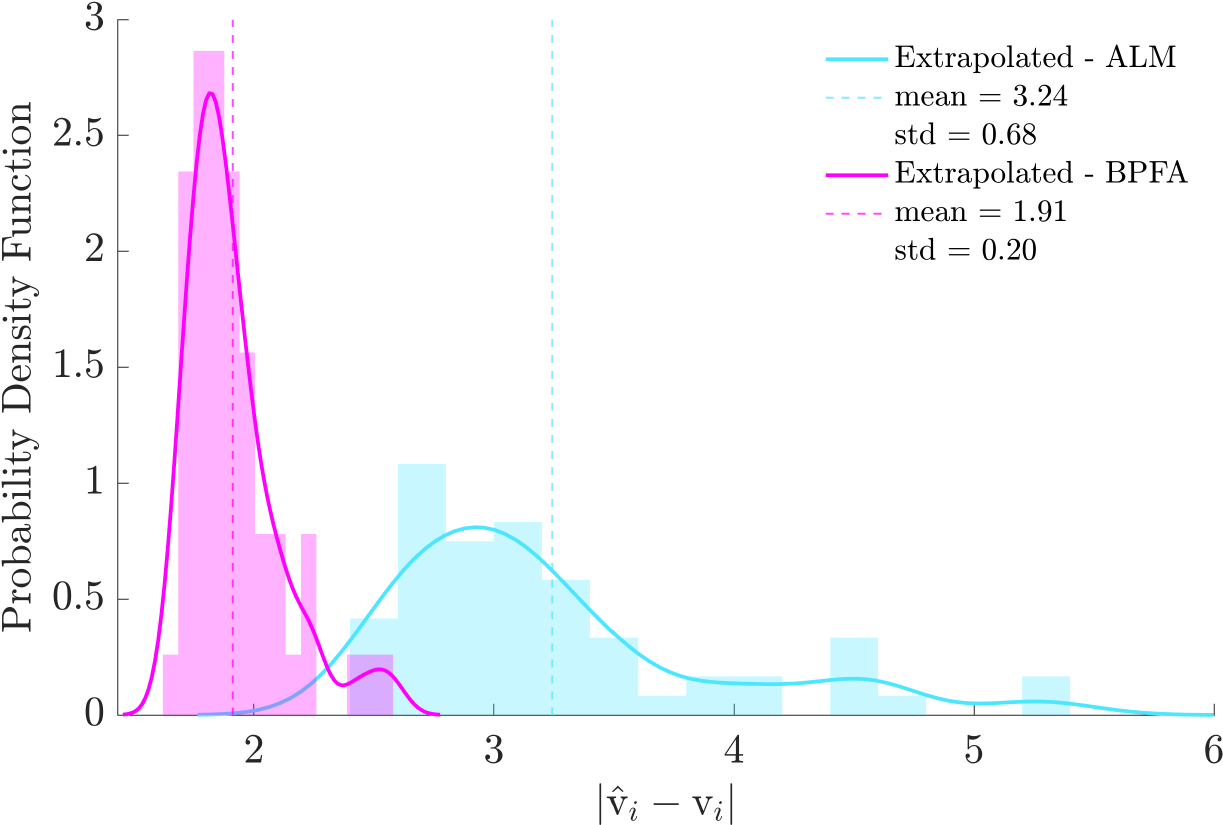}
    \caption{Distribution of the average reconstruction error across the ensemble MCS realizations at all 60 grid points of Fig.~\ref{Schematic_2D} using the ALM algorithm and the BPFA approach.}\label{error_map_2D}
\end{figure}

\begin{figure}[!ht]
    \centering    \includegraphics[width=0.7\textwidth]{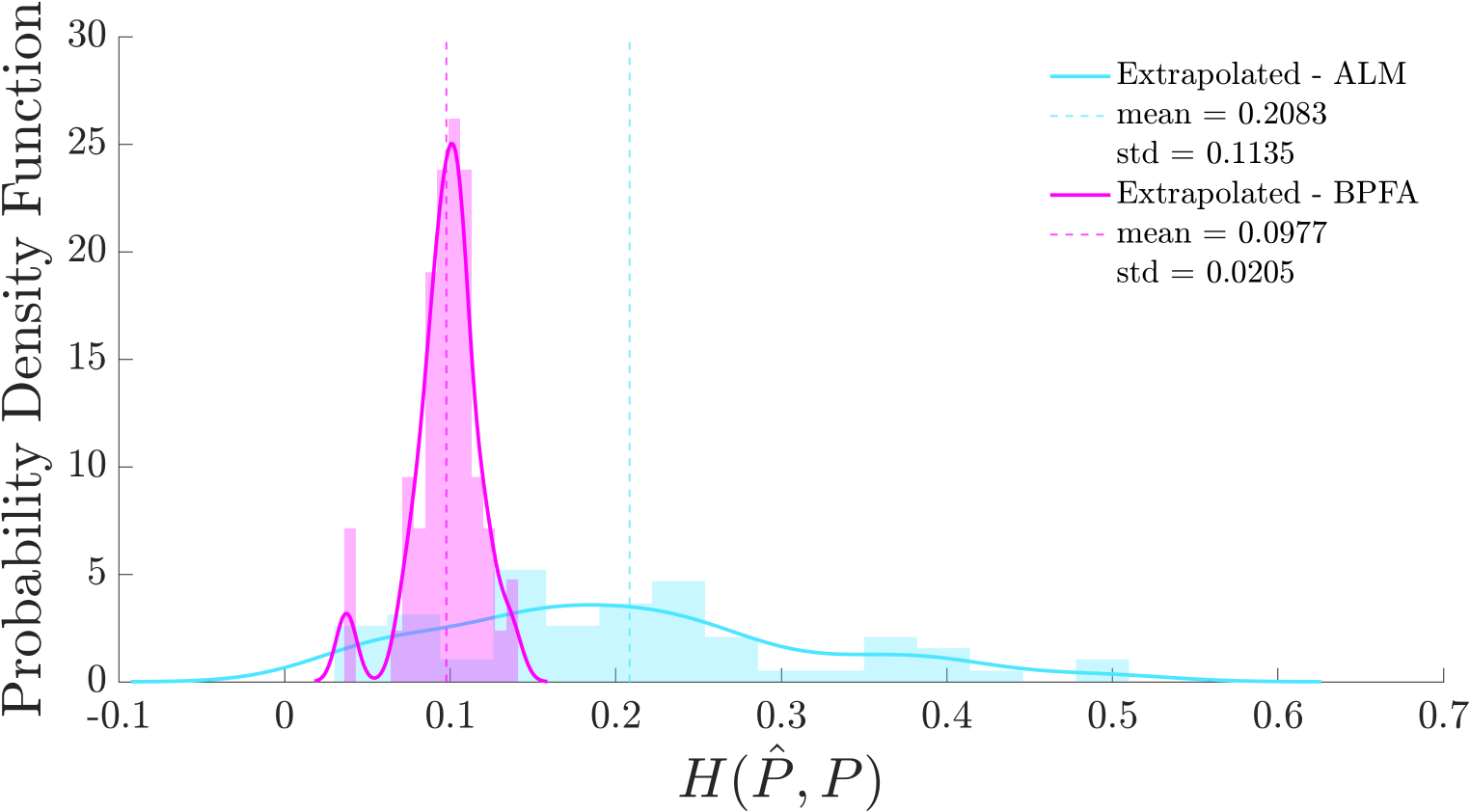}\caption{Distribution of the Heilinger distance in Eq.~\eqref{Heilinger} considering all 60 missing grid points of Fig.~\ref{Schematic_2D} for the ALM algorithm and the BPFA approach.}\label{hist_error_map_2D}
\end{figure}

A detailed analysis of the reconstruction error is performed by comparing the distribution of the average $\ell_1$-error, \(\|\mathrm{\hat{v}_i}-\mathrm{v_i}\|_1\), across all MCS realizations between the true record \(\mathrm{v_i}\) and the extrapolated predictions \(\mathrm{\hat{v}_i}\) at all 60 grid points of Fig.~\ref{Schematic_2D}. The results, shown in Fig.~\ref{error_map_2D}, demonstrate the efficacy of the BPFA method which yields a considerably lower mean and variance error. Further, the Heilinger distance denoted by 
\begin{equation}\label{Heilinger}
    H(\hat{P}, P) = \frac{1}{\sqrt{2}} \left\| \sqrt{\hat{P}} - \sqrt{P} \right\|_2
\end{equation}
is used for comparing the estimated distributions of the extrapolated and original time histories, \(\hat{P}\) and  \(P\), respectively, at each of the missing locations obtained for all MCS samples. The distribution of these discrepancies is visualized in Fig. \ref{hist_error_map_2D} where the BPFA demonstrates lower mean error and lower error variance compared to the ALM. Thus, it can be argued that BPFA extrapolates records in a more statistically consistent and reliable manner compared to the ALM.


\begin{figure}[!ht]
    \centering    \includegraphics[width=0.7\textwidth]{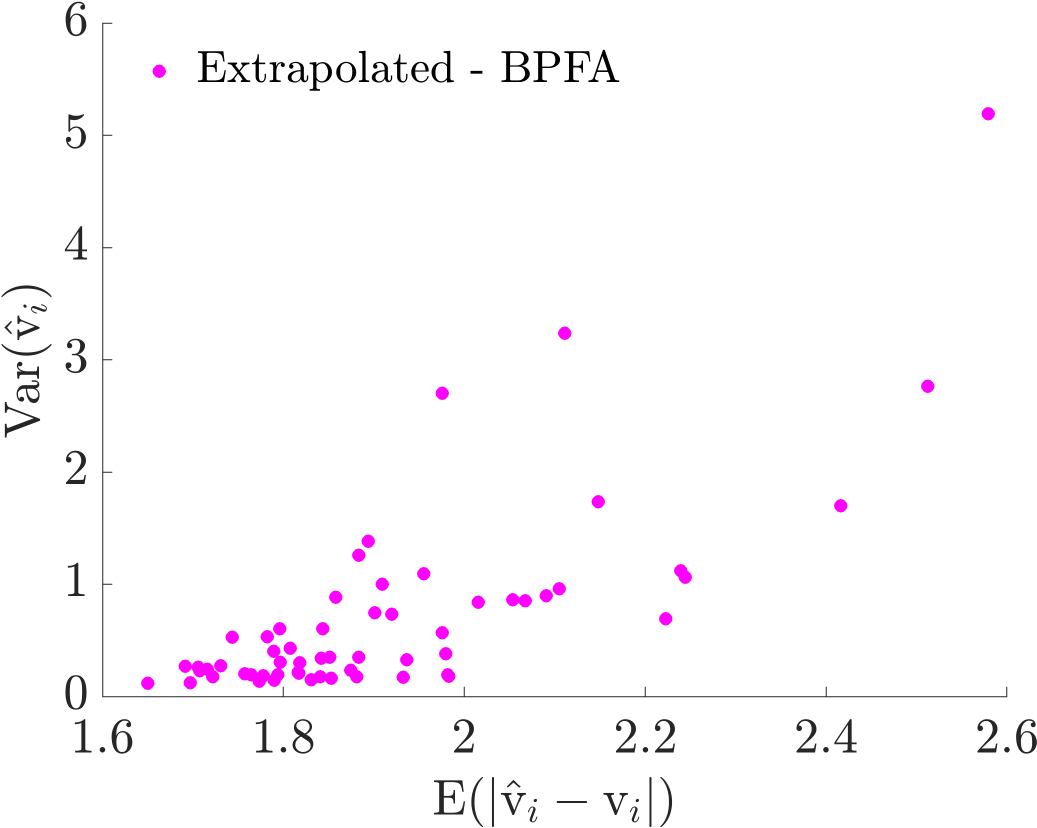}
    \caption{Averaged time-history reconstruction error versus the posterior variance from the BPFA algorithm for all 60 grid points of Fig.~\ref{Schematic_2D} with missing records.}\label{error_vs_variance_2D}
\end{figure}

Finally, Fig.~\ref{error_vs_variance_2D} shows the time history reconstruction error versus the posterior variance at all of the 60 missing grid points of Fig.~\ref{Schematic_2D} for the BPFA algorithm. It is observed that the variance is well correlated with the reconstruction error. Thus, the BPFA method yields reasonable uncertainty estimates in the presence of missing data and could be used as a reliable predictive tool in wind engineering applications. 

\newpage

\section{Wind field extrapolation in three spatial dimensions}
In this section, the efficacy of the sparse representation methodologies is assessed in three spatial dimensions in conjunction with wind field data obtained by BLWT experiments. The data acquisition process is briefly detailed in the following section. 

\subsection{Boundary Layer Wind Tunnel (BLWT) data acquisition and processing}\label{section31}
\begin{figure}
    \centering
    \includegraphics[width=\textwidth]{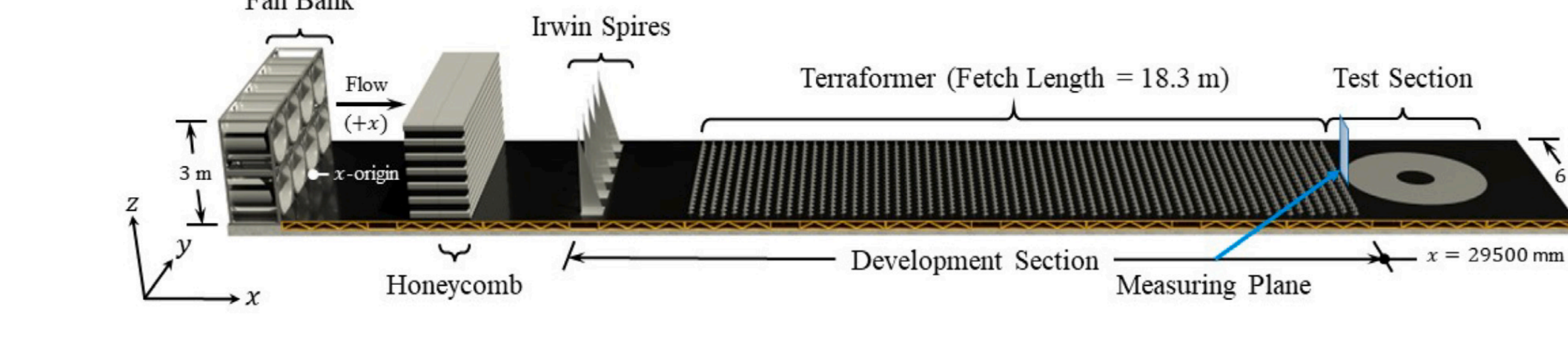}\caption{Schematic of the University of Florida BLWT. The fan bank of vaneaxial fants drives wind down the tunnel and over the Terraformer roughness element grid to produce boundary layer-like wind flow profiles.}\label{BLWT}
\end{figure}

\begin{figure}
    \centering 
    \includegraphics[width=0.8\textwidth]{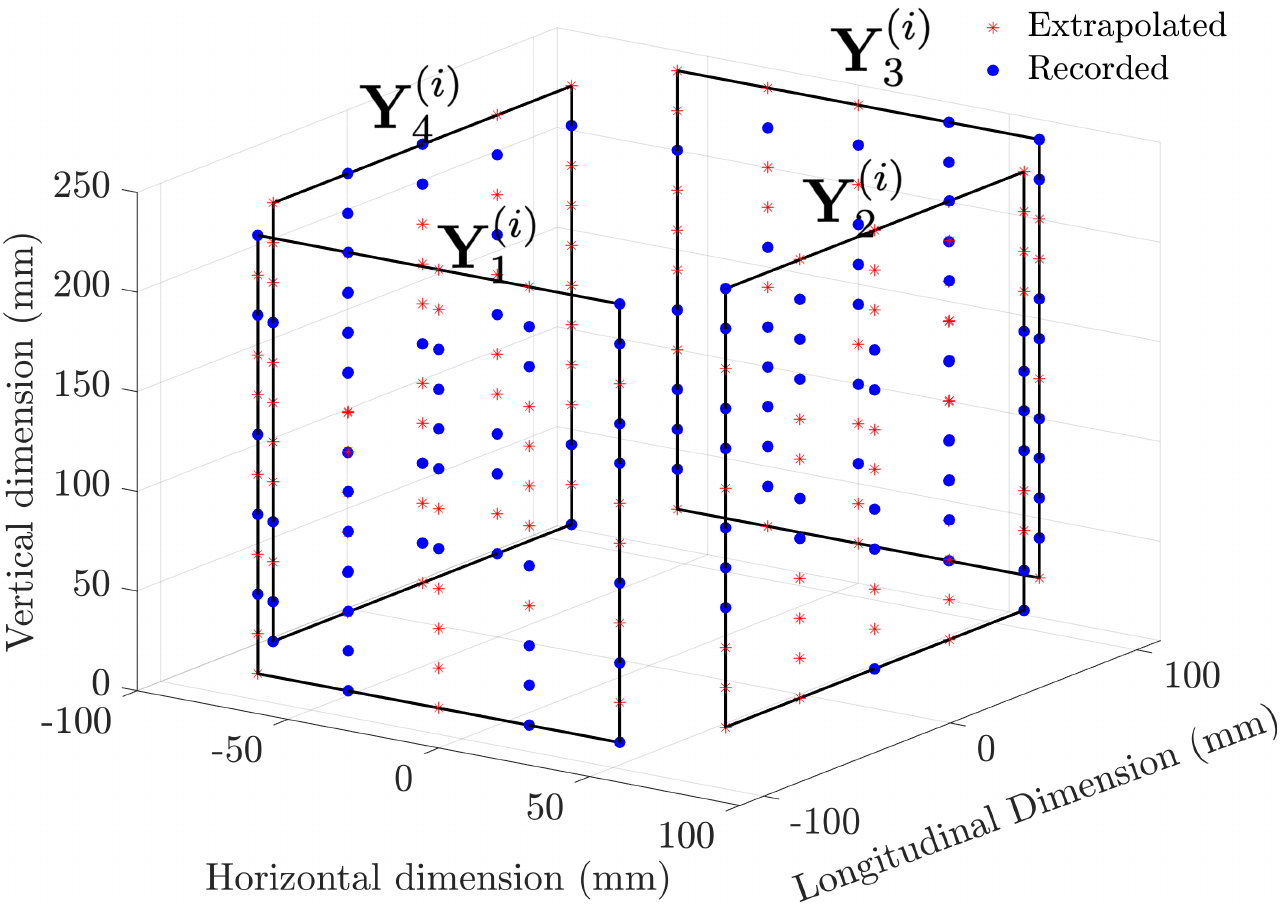}
    \caption{Schematic showing the faces of the model structure with pressure taps (blue circles and red stars) distributed across all four faces. Each face has 60 pressure taps that record the x, y, and z components of the wind pressure. For the purposes of extrapolation, data collected at the \(120\) red points are ignored and extrapolated from recordings at the \(120\) blue points.
    }
    \label{Schematic_3D}
\end{figure}

The data were collected during a series of studies to analyze non-Gaussian wind effects on structures conducted at the University of Florida BLWT facility~\cite{shields2023active,ojeda2025modulating}. The data are publicly available on the DesignSafe-CI~\cite{ojeda2025designsafe}. The experiments involve wind blown along a roughly \(40 m\) channel by 8 vaneaxial fans with a set of automatically configurable mechanical roughness elements (the Terraformer) generating boundary layer-like turbulence as illustrated in Figure~\ref{BLWT}. 
Located at the test section, a \(150 \ mm \times 225 \ mm \times 250 \ mm\) scale model structure is placed in the wind field with pressure taps on all four faces (and the roof, which is not studied here) as shown in Figure~\ref{Schematic_3D}. 

The available measurements consist of wind pressures, herein denoted as \(\mathrm{p}_i\), recorded at the \(240\) pressure tap locations on the faces of the experimental structure. 
% acquired during bluff body experiments. 
A total of \(10\) distinct experiments are considered, with each sampled at a frequency of \(f_s = 625 Hz\) resulting in \(N=6554\) points over a total time of \(T_0=104.85s\). The measurement probes provide data in \(240\) three dimensional points out of which \(120\) are considered randomly missing resulting to \(50\%\) missing data. A schematic representation of the spatial data acquisition pattern is shown in Fig.~\ref{Schematic_3D}. The variables \(\mathbf{Y}_1^{(i)},\mathbf{Y}_2^{(i)},\mathbf{Y}_3^{(i)},\mathbf{Y}_4^{(i)}\) denote the sampling pattern at each face of the structure at each time instant. 

A matrix reshape is adopted in which the measurement grid is concatenated in the following form: \(\mathbf{Y} = \left[\mathbf{Y}^{(1)}; \mathbf{Y}^{(2)}; \mathbf{Y}^{(3)}; \mathbf{Y}^{(4)}\right]\). This leads to a rectangular data matrix \( \asb{Y} \in \mathbb{R}^{M_x \times M_z \times N}\) where \(M_x = 12, M_y = 20, N = 6554\). 

\begin{figure}
    \centering    
    \includegraphics[width=\textwidth]{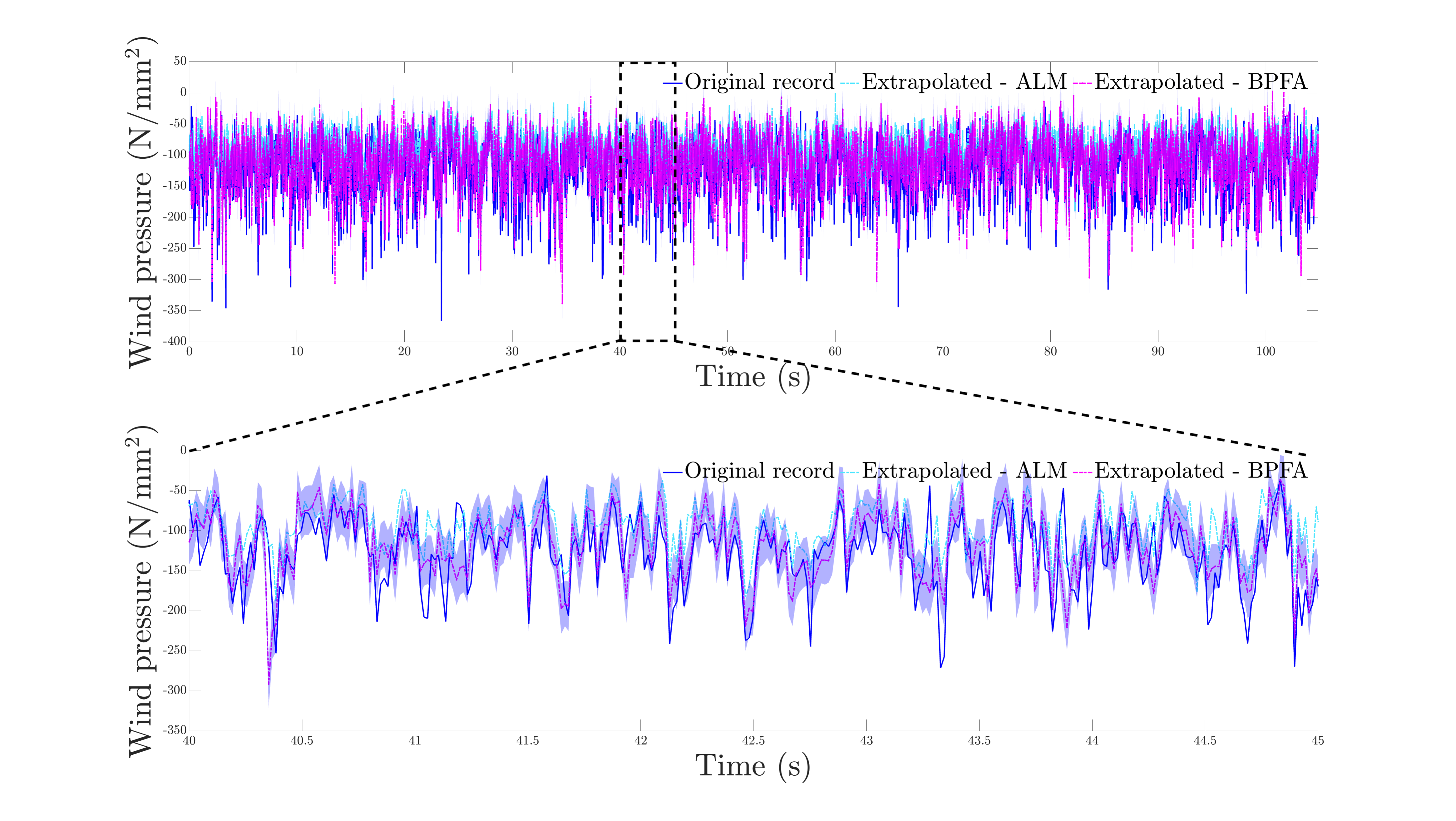}
    \caption{Extrapolated time history at point \((-75,80,240)\), as shown in Fig.~\ref{Schematic_3D}. Top figure: Full record. Bottom figure: 5 second segment of the record showing detailed fluctuations.}
    \label{Sample_comparison_(-75_80_240)}
\end{figure}

\begin{figure}
    \centering
    \includegraphics[width=\textwidth]{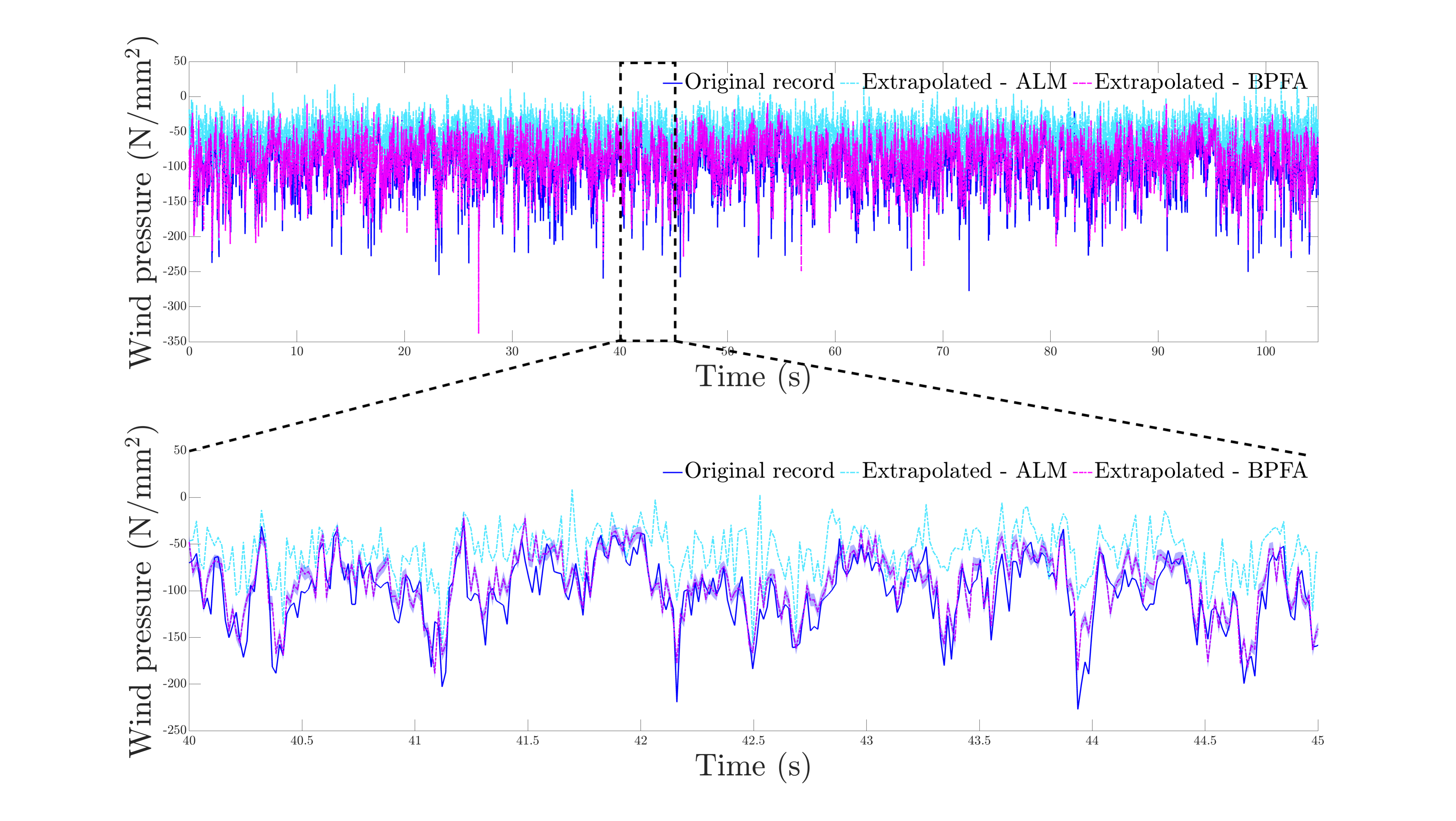}
    \caption{Extrapolated time history at point \((75,-40,160)\), as shown in Fig.~\ref{Schematic_3D}. Top figure: Full record. Bottom figure: 5 second segment of the record showing detailed fluctuations.}
    \label{Sample_comparison_(75_-40_160)}
\end{figure}

\begin{figure}
    \centering  
    \includegraphics[width=0.7\textwidth]{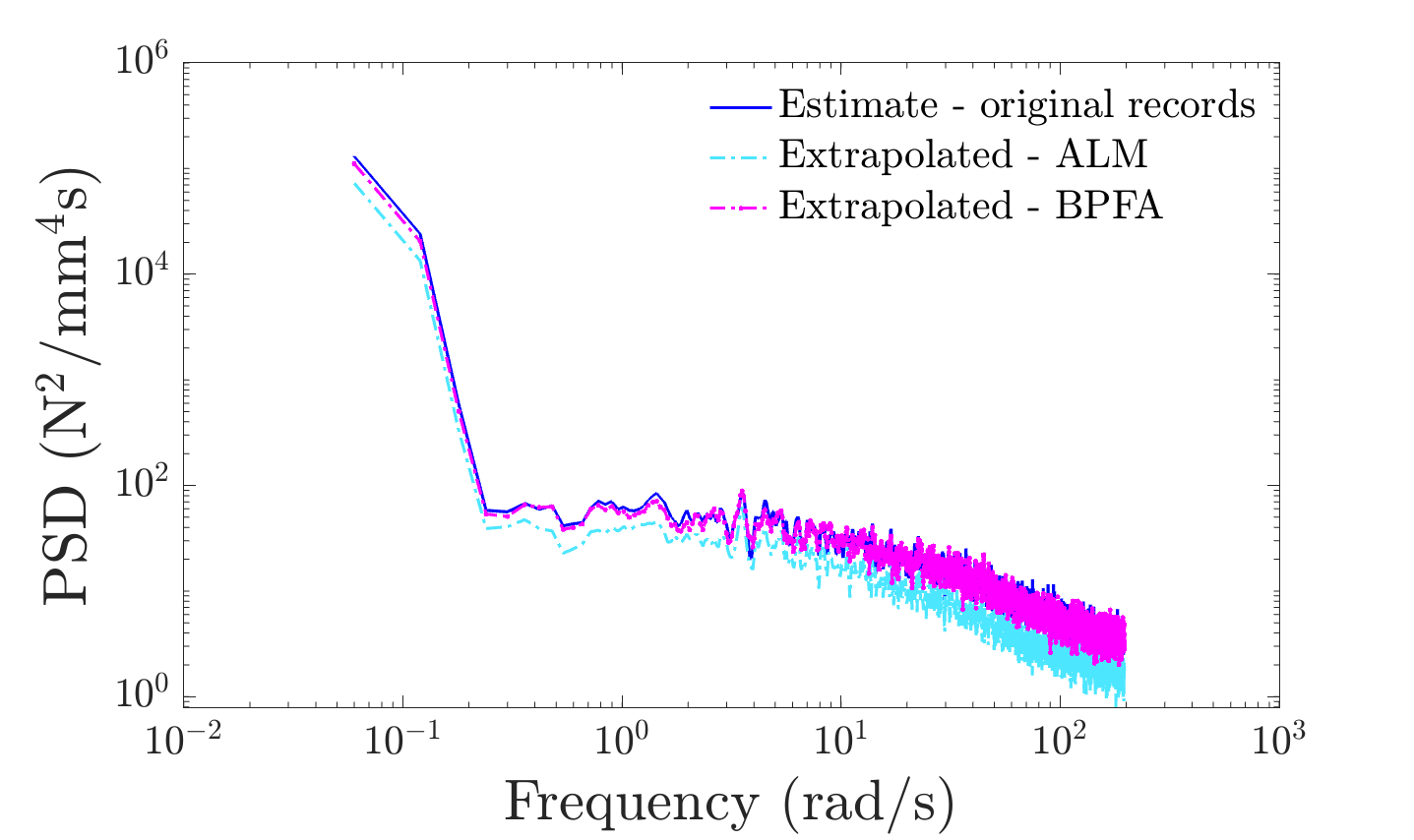}
    \caption{Comparison of the estimated PSD at point \((-75,80,240)\) as shown in Fig.~\ref{Schematic_3D} based on the ensemble average of the ALM-based algorithm and the BPFA-based approach.}
    \label{PSD_-75_80_240}
\end{figure}

\begin{figure}
    \centering  
    \includegraphics[width=0.7\textwidth]{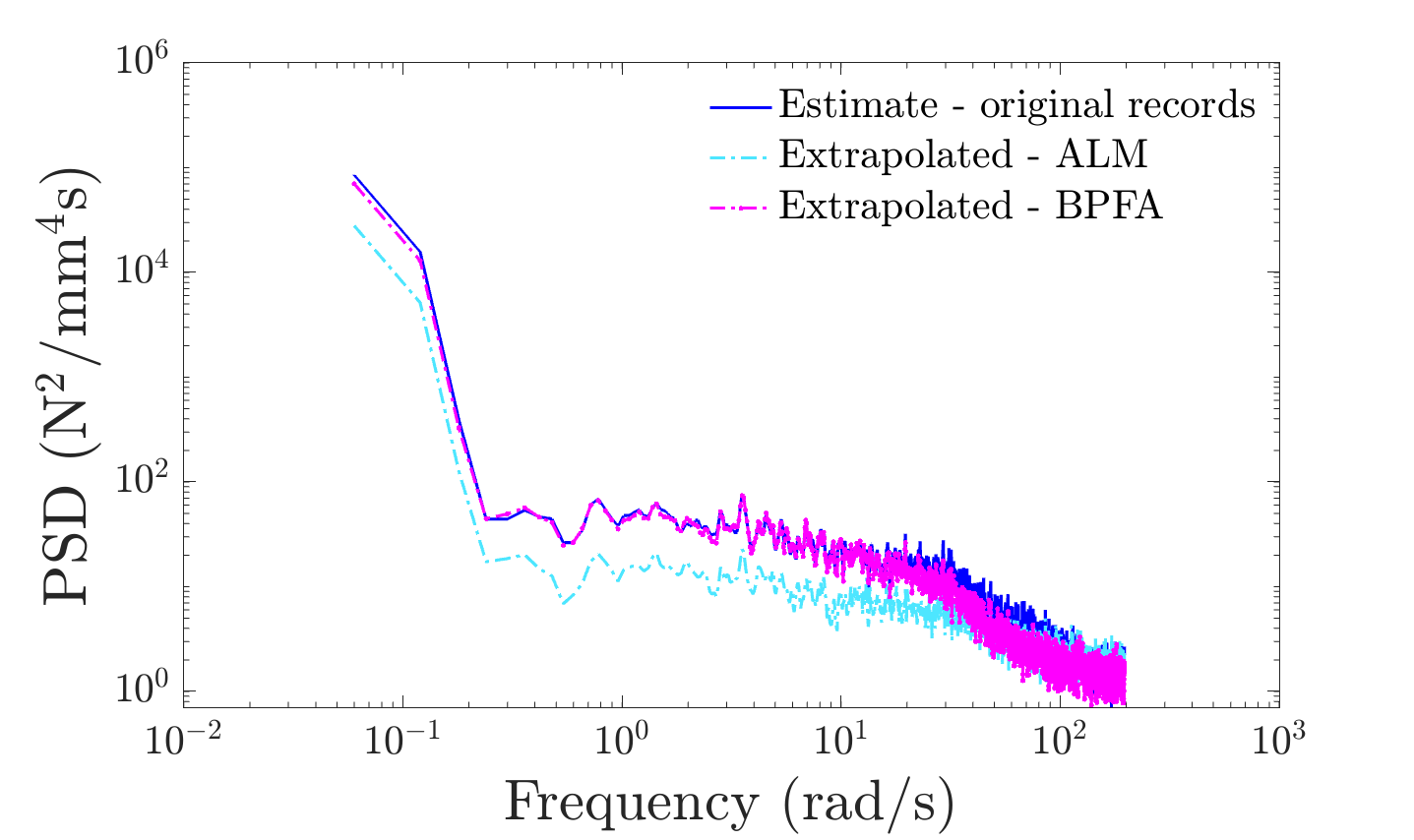}
    \caption{Comparison of the estimated PSD at point \((75,-40,160)\) as shown in Fig.~\ref{Schematic_3D} based on the ensemble average of the ALM-based algorithm and the BPFA-based approach.}
    \label{PSD_75_-40_160}
\end{figure}

\begin{figure}
    \centering
    \includegraphics[width=0.7\textwidth]{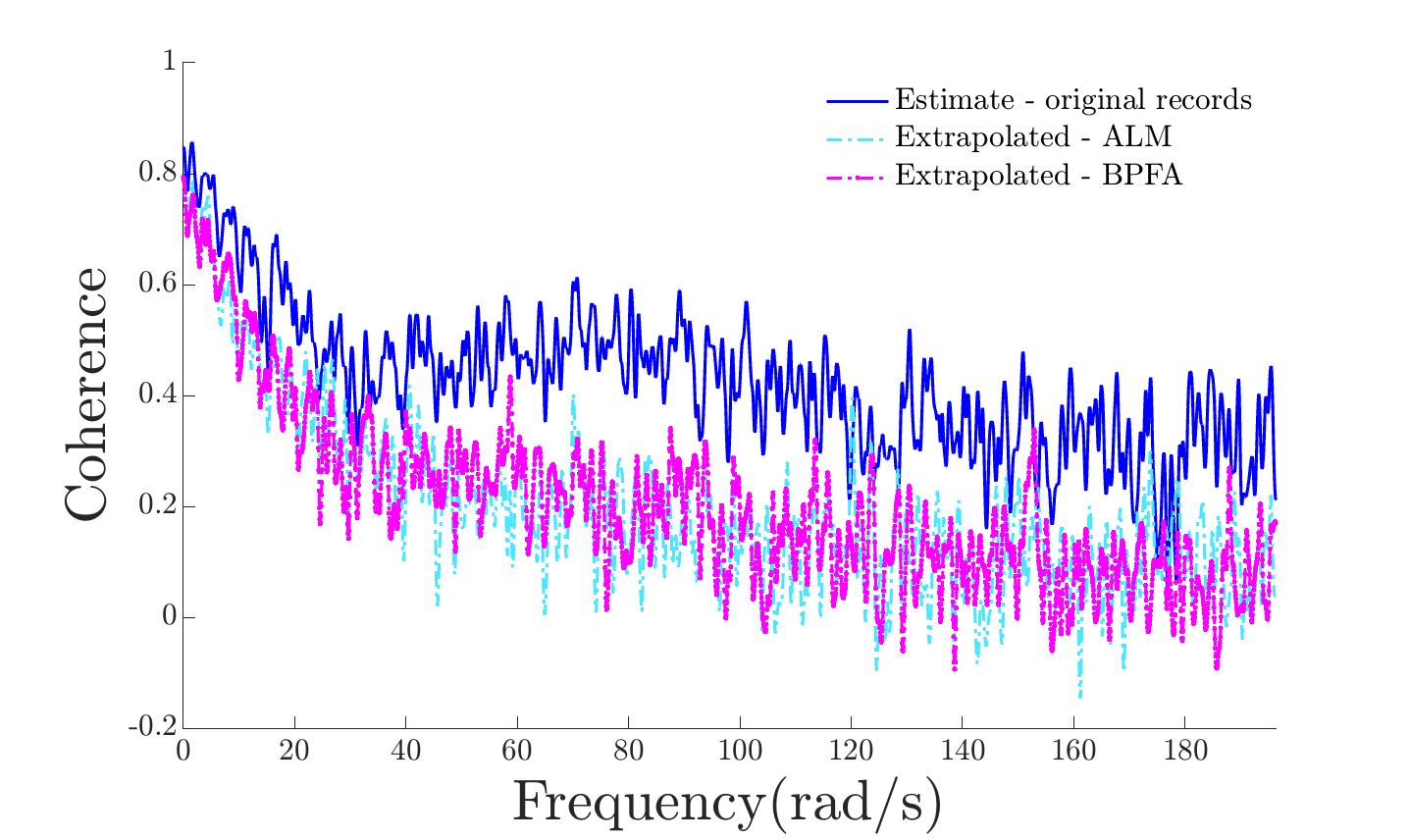}
    \caption{Comparison of the estimated cross-coherence between points \((-75,80,240)\) and \((75,-40,160)\) as shown in Fig.~\ref{Schematic_3D} based on the ensemble average of the ALM-based algorithm and the BPFA-based approach.}
    \label{Coh_comparison_(-75_80_240_75_-40_160)}
\end{figure}

Next, the optimization framework shown in Fig.~\ref{Schematic_2D} is employed for reconstructing the records at each missing index and for each sample realization. The same hyperparameters have been used for both algorithms as in section \ref{numerical_example_1}. Results pertaining to two representative time histories are shown in Fig.~\ref{Sample_comparison_(-75_80_240)} and Fig.~\ref{Sample_comparison_(75_-40_160)}. It can be seen that the BPFA method yields enhanced accuracy in reconstructing the time-histories compared to the ALM-based approach. The corresponding PSDs are shown in Fig.~\ref{PSD_-75_80_240} and Fig.~\ref{PSD_75_-40_160}, for the two time-histories, where the BPFA method furnishes enhanced results in capturing both the peak power and the overall PSD characteristics. A similar trend is observed in Fig.~\ref{Coh_comparison_(-75_80_240_75_-40_160)} which shows the associated cross-coherence between the two points.

\begin{figure}
    \centering
    \includegraphics[width=0.6\textwidth]{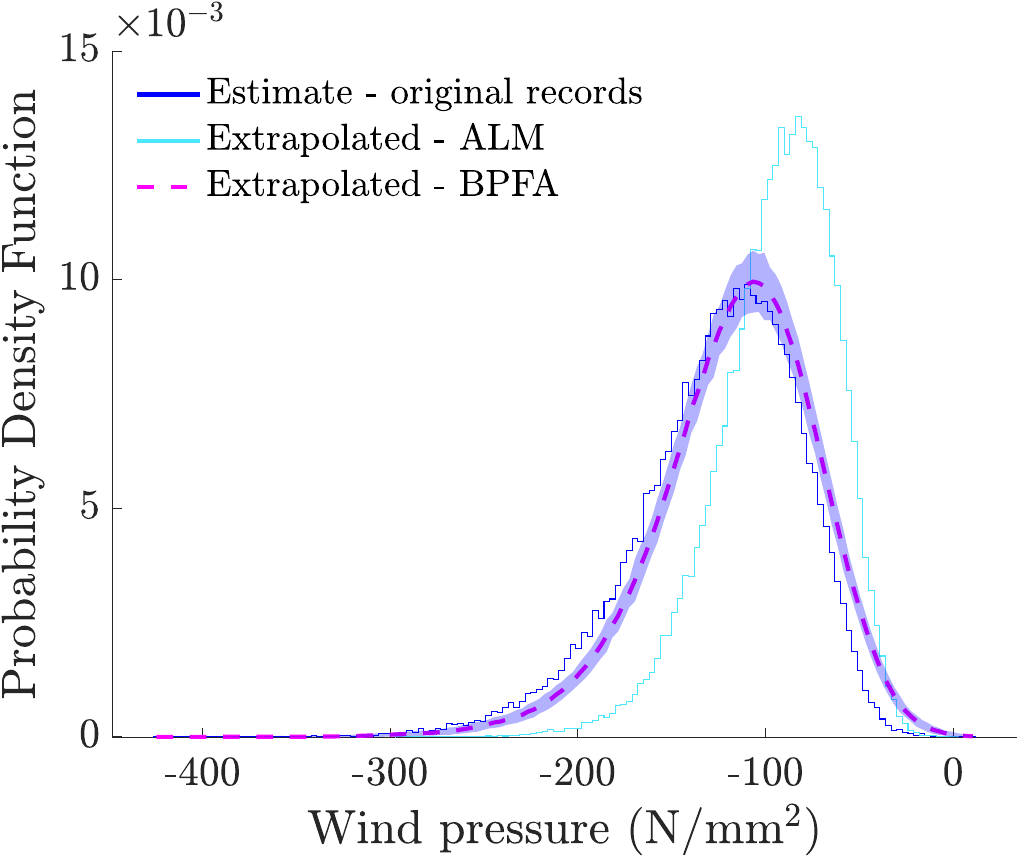}
    \caption{Estimated PDF of the extrapolated time-histories at point \((-75,80,240)\), as shown in Fig.~\ref{Schematic_3D}, based on the ensemble of the ALM-based algorithm and the BPFA-based approach.}
    \label{Hist_comparison_(-75_80_240)}
\end{figure}
\begin{figure}
    \centering
    \includegraphics[width=0.6\textwidth]{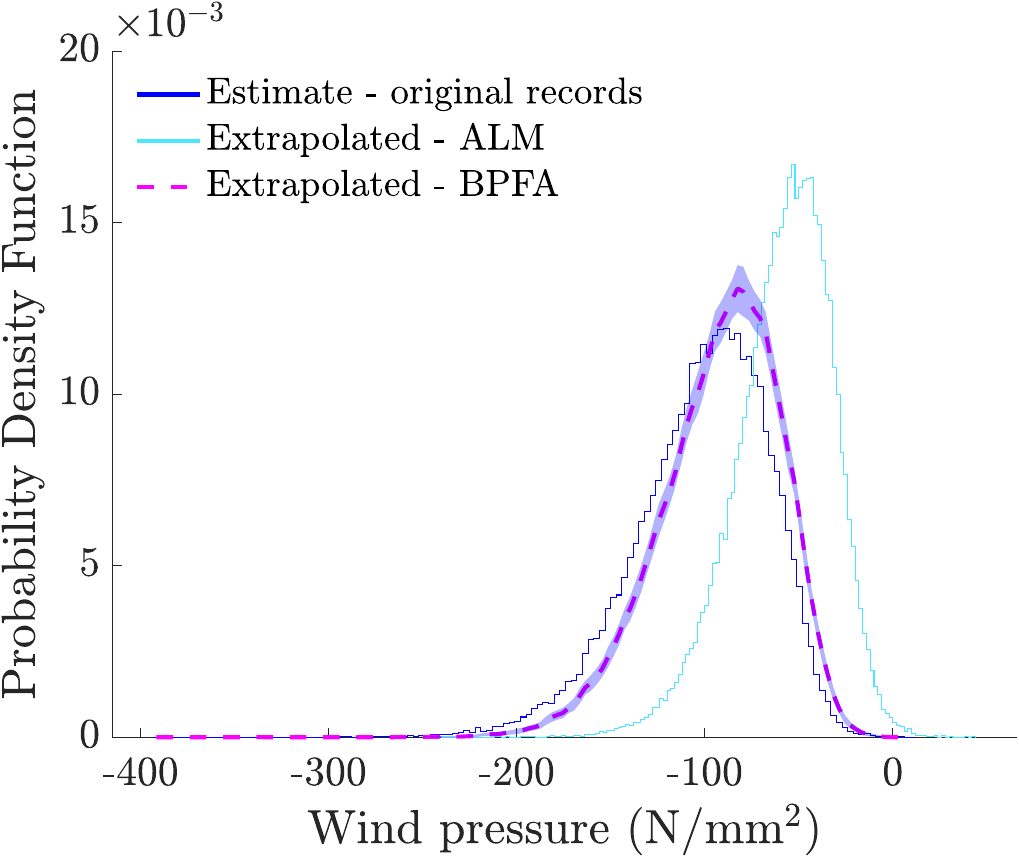}
    \caption{Estimated PDF of the extrapolated time-histories at point \((75,-40,160)\), as shown in Fig.~\ref{Schematic_3D}, based on the ensemble of the ALM-based algorithm and the BPFA-based approach.}
    \label{Hist_comparison_(75_-40_160)}
\end{figure}
\begin{figure}[!ht]
    \centering
    \includegraphics[width=0.7\textwidth]{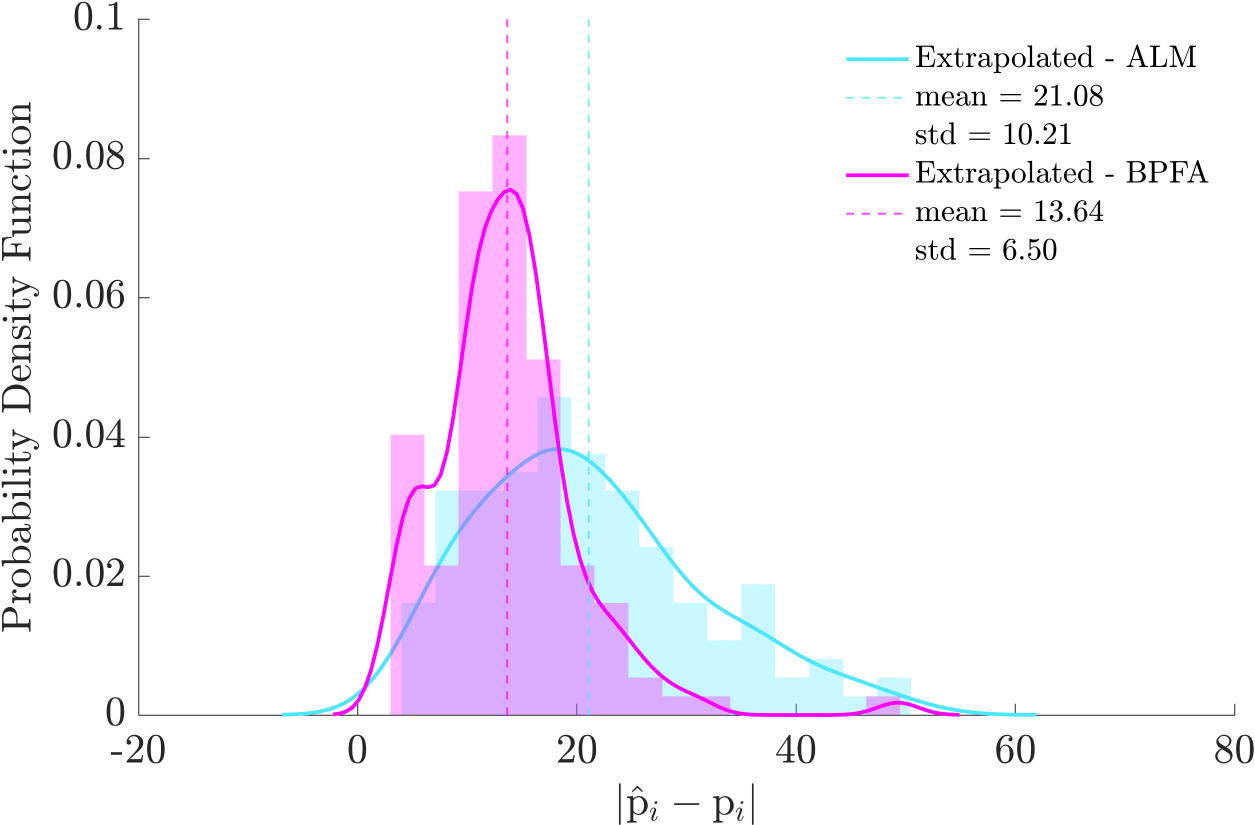}
    \caption{Distribution of the average reconstruction error across the ensemble MCS realizations at all 120 missing points of Fig.~\ref{Schematic_3D} using the ALM algorithm and the BPFA approach.}\label{error_map_3D}
\end{figure}
\begin{figure}[!ht]
    \centering
    \includegraphics[width=0.7\textwidth]{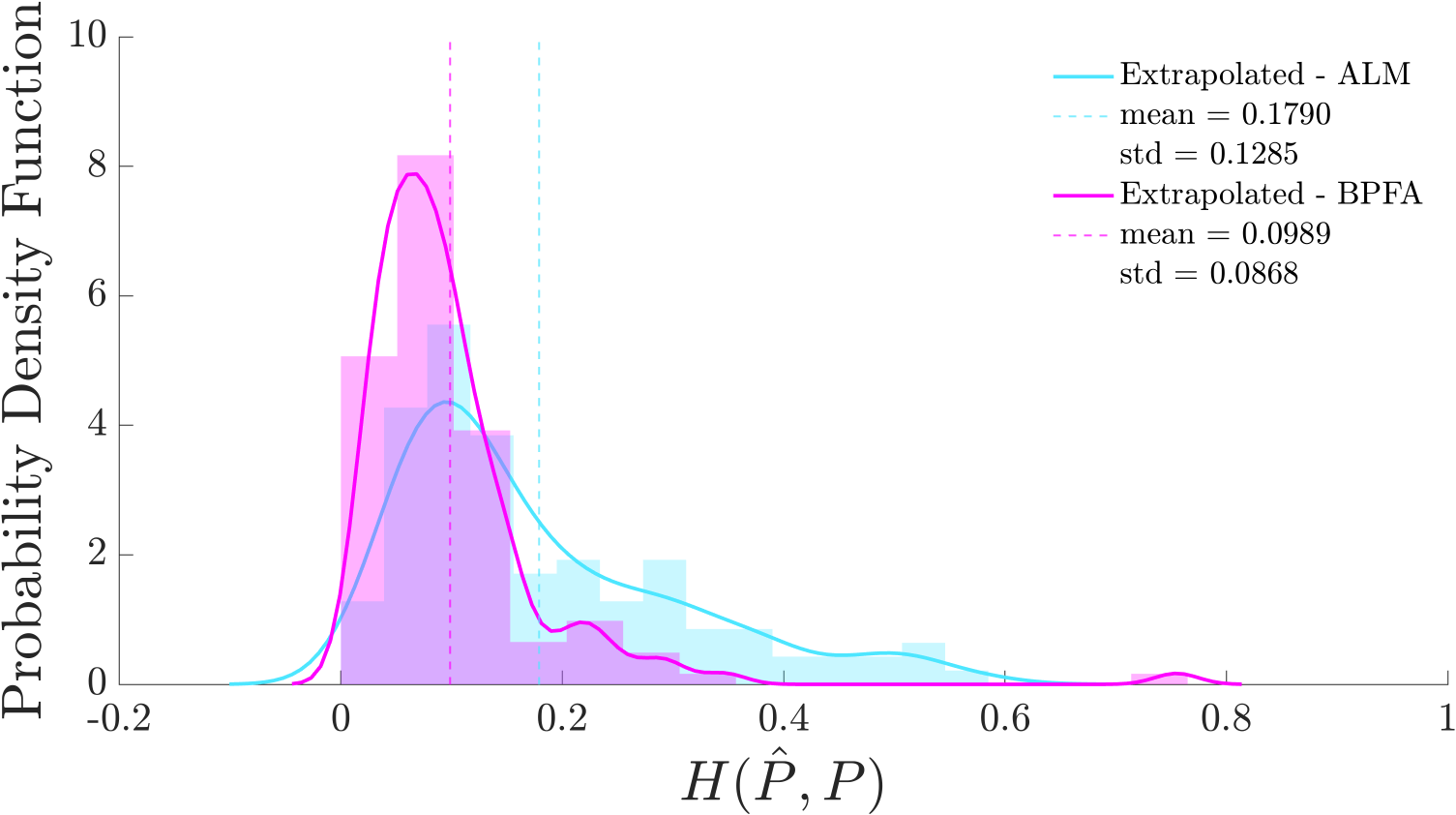}\caption{Distribution of the Heilinger distance in Eq.~\eqref{Heilinger} considering all 120 missing grid points of Fig.~\ref{Schematic_3D} for the ALM algorithm and the BPFA approach.}\label{hist_error_map_3D}
\end{figure}

Further, the estimated PDFs of the reconstructed pressure measurements are shown in Fig.~\ref{Hist_comparison_(-75_80_240)} and Fig.~\ref{Hist_comparison_(75_-40_160)}, respectively, considering all MCS realizations. It is worth noting that, although both methods provide relatively accurate time-history reconstructions, the BPFA method estimates the non-Gaussian PDF markedly better. This is also corroborated by the results in Fig.~\ref{error_map_3D} in which the estimated PDF of the averaged reconstruction error \(\|\mathrm{\hat{p}_i}-\mathrm{p_i}\|_1\) is obtained across all MCS realizations and for all missing locations. Similarly, in Fig.~\ref{hist_error_map_3D}, the Heilinger distance in Eq.~\eqref{Heilinger} is used to compare the error between the estimated distributions of the extrapolated and original time histories, \(\hat{P}\) and  \(P\), respectively, at each of the missing locations and for all MCS samples. It is seen that the average reconstruction error as well as the distribution discrepancy is lower and more narrowly concentrated in the case of the BPFA algorithm as compared to the ALM.

\begin{figure}
    \centering
    \includegraphics[width=0.7\textwidth]{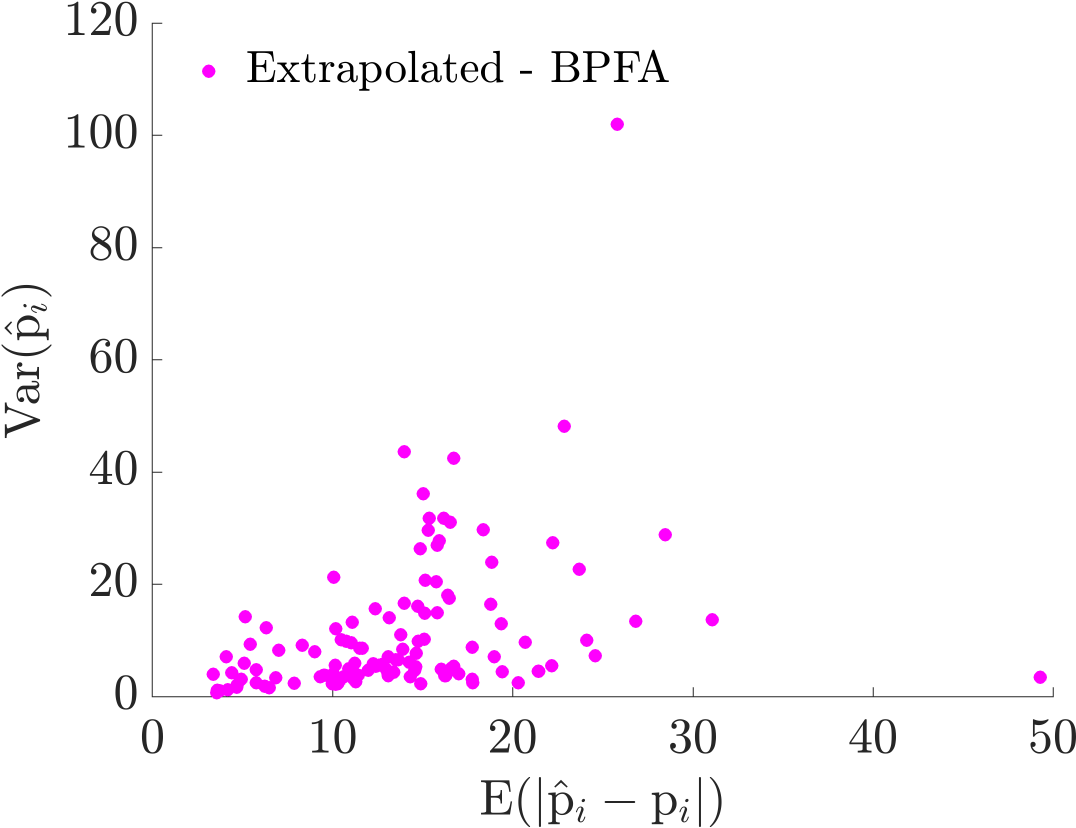}
    \caption{Averaged time-history reconstruction error vis-à-vis the posterior variance of the unknown grid points of Fig.~\ref{Schematic_3D}.}\label{error_vs_variance_3D}
\end{figure}

Further, in the case of the BPFA algorithm, the predictive variance is plotted against the MCS averaged reconstruction error in Fig.~\ref{error_vs_variance_3D}. It can be argued that the BPFA approach yields an overall lower reconstruction error across all metrics while also accompanied by reasonable uncertainty estimates. Thus, it is a promising tool which can potentially reduce significantly the number of sensors in BLWT testing.


\section{Concluding remarks}
A novel methodology has been developed in this paper for stochastic wind field reconstruction and extrapolation based on a limited amount of measured data. The methodology exhibits significant advantages compared to alternative state-of-the-art approaches based, for instance, on CS and low-rank matrix concepts and tools. Specifically, while CS approaches have demonstrated success in reconstructing signals based on limited/incomplete measurements, their dependence on predefined basis expansions renders their applicability limited in a high-dimensional setting involving data of arbitrary form. Further, although low-rank matrix methods are comparatively more  scalable, they generally fail to sufficiently capture temporal correlations across wind-time histories, and do not provide uncertainty estimates in the predictions. To address these challenges, the herein developed methodology has exploited the concept of Bayesian dictionary learning for determining, in an adaptive manner, a low-dimensional representation of the stochastic wind field, where the associated expansion coefficients are inferred from the observed data. Compared to earlier efforts by some of the authors based on a standard CS framework \cite{pasparakis2022wind}, the proposed methodology has demonstrated significant improvements in terms of extrapolation accuracy, particularly in scenarios with large amounts of missing data. Also, it has shown to be computationally tractable even in four-dimensional (3D + time) domains. Further, the methodology yields uncertainty estimates that are well correlated with extrapolation accuracy, learned through a hierarchical model for the prediction error. This aspect demonstrates the potential of the methodology as a reliable predictive tool. Various numerical examples have been considered relating to both simulated and BLWT data. The methodology has exhibited a satisfactory degree of reliability in data extrapolation, even in cases of relatively large distances, and of non-Gaussian data. The latter highlights also the practical merit of the methodology since the required number of sensors in wind engineering applications can be potentially reduced.
\FloatBarrier
\newpage
\newpage
\appendix
\section{Gibbs sampling updates for Eq.~\eqref{BP_posterior} ins Section \ref{DL_section}}{\label{appendix}}
The posterior distribution in Eq.~\eqref{BP_posterior} is inferred using Gibbs sampling steps where each variable is iteratively updated by sampling the conditional distribution of the remaining variables. These steps can be found in \cite{zhou2011nonparametric} and are presented below for completeness. 

The observation matrix \(\mathbf{\Sigma}_i\) in Eq.~\eqref{CS_BPFA} is constructed by removing the rows of the identity matrix which correspond to the missing data locations. In this regard, \(\mathbf{\Sigma}_i^\top \mathbf{\Sigma}_i\) is a sparse identity matrix with the property
\begin{equation}
\mathbf{\Sigma}_i \mathbf{\Sigma}_i^\top = \mathbf{I}_{\|\mathbf{\Sigma}_i\|_0}
\end{equation}
The sampling scheme starts by considering the conditional distribution for the columns of \(\mathbf{D}\) given by
\begin{equation}
    p(\mathbf{d}_k | - ) \propto \prod_{i=1}^{n_{total}} \mathcal{N}\left(\mathbf{y}_i ; \mathbf{\Sigma}_i \mathbf{D}(\mathbf{s}_i \odot \mathbf{z}_i), \gamma_\epsilon^{-1} \mathbf{I}_{\|\mathbf{\Sigma}_i\|_0} \right) \mathcal{N}(\mathbf{d}_k ; \mathbf{0}, P^{-1} \mathbf{I}_P)
\end{equation}
Using the properties of the conjugate-exponential distributions, the \(\mathbf{d}_k\) atoms are sampled from a normal distribution as
\begin{equation}
    p(\mathbf{d}_k | - ) \sim \mathcal{N}(\boldsymbol{\mu}_{d_k}, \mathbf{\Sigma}_{d_k})
    \end{equation}
with mean and covariance given by
\begin{equation}
    \boldsymbol{\mu}_{\mathbf{d_k}} = \gamma_\epsilon \mathbf{\Sigma}_{\mathbf{d_k}} \sum_{i=1}^{n_{total}} z_{ik} s_{ik}
    \tilde{\mathbf{x}}_i^k
    \end{equation}
\begin{equation}
    \mathbf{\Sigma}_{d_k} = \left(P\mathbf{I} + \gamma_\epsilon \sum_{i=1}^{n_{total}} z_{ik}^2 s_{ik}^2 \mathbf{\Sigma}_i^\top \mathbf{\Sigma}_i \right)^{-1}
    \end{equation}
respectively, where \(\tilde{\mathbf{x}}_i^k\) is given by
\begin{equation}
    \tilde{\mathbf{x}}_i^{-k} = \mathbf{\Sigma}_i^\top \mathbf{y}_i 
    - \mathbf{\Sigma}_i^\top \mathbf{\Sigma}_i \mathbf{D}(\mathbf{s}_i \odot \mathbf{z}_i) 
    + \mathbf{\Sigma}_i^\top \mathbf{\Sigma}_i \mathbf{d}_k (s_{ik} \odot z_{ik})
\end{equation}
Next, the binary vector \(\mathbf{z}_i = [z_{i1}, z_{i2}, \dots, z_{iK}]\) is sampled from the distribution
\begin{equation}
    p(z_{ik} | - ) \propto \mathcal{N}\left(\mathbf{y}_i ; \mathbf{\Sigma}_i \mathbf{D}(\mathbf{s}_i \odot \mathbf{z}_i), \gamma_\epsilon^{-1} \mathbf{I}_{\|\mathbf{\Sigma}_i\|_0} \right) \prod_{k=1}^K \text{Bernoulli}(z_{ik}; \pi_k)
    \end{equation}
Considering that the posterior probability of the event \(z_{ik}=1\) is proportional to
\begin{equation}
    p_1 = \pi_k \exp\left[ -\frac{\gamma_\epsilon}{2} \left( s_{ik}^2 \mathbf{d}_k^\top \mathbf{\Sigma}_i^\top \mathbf{\Sigma}_i \mathbf{d}_k - 2 s_{ik} \mathbf{d}_k^\top \tilde{\mathbf{x}}_i^{-k} \right) \right]
\end{equation}
and that the posterior probability of the event \(z_{ik}=0\) is
\begin{equation}
    p_0 = 1 - \pi_k
\end{equation}
yields the following sampling distribution for \(z_{ik}\), i.e., 
\begin{equation}
    z_{ik} \sim \text{Bernoulli} \left( \frac{p_1}{p_0 + p_1} \right)
\end{equation}
Further, the distribution of the coefficient vector \(\mathbf{s}_k := [s_{1k}, s_{2k}, \dots, s_{Nk}]\) is proportional to the distribution
\begin{equation}
    p(s_{ik} | - ) \propto \mathcal{N}\left(\mathbf{y}_i ; \mathbf{\Sigma}_i \mathbf{D}(\mathbf{s}_i \odot \mathbf{z}_i), \gamma_\epsilon^{-1} \mathbf{I}_{\|\mathbf{\Sigma}_i\|_0} \right) \mathcal{N}(\mathbf{s_{i}}; 0, \gamma_s^{-1}\mathbf{I}_K)
\end{equation}
which can be shown to follow the Gaussian distribution
\begin{equation}
    p(s_{ik}|-) \sim \mathcal{N}(\mu_{s_{ik}}, \Sigma_{s_{ik}})
\end{equation}
with mean and covariance given by
\begin{equation}
    \mu_{s_{ik}} = \gamma_\epsilon \Sigma_{s_{ik}} z_{ik} \mathbf{d}_k^\top \mathbf{\Sigma}_i^\top \mathbf{\Sigma}_i \tilde{\mathbf{x}}_i^{-k}
    \end{equation}
\begin{equation}
    \Sigma_{s_{ik}} = \left( \gamma_s + \gamma_\epsilon z_{ik}^2 \mathbf{d}_k^\top \mathbf{\Sigma}_i^\top \mathbf{\Sigma}_i \mathbf{d}_k \right)^{-1}
    \end{equation}
respectively. The distribution of the hyperparameter \(\pi_k\) associated with the binary vector \(\mathbf{z}_i\) is proportional to a Beta-Bernoulli distribution in the form
\begin{equation}
    p(\pi_k \mid -) \propto \text{Beta}\left(\pi_k ; \frac{a}{K}, \frac{b(K-1)}{K} \right) \prod_{i=1}^{n_{total}} \text{Bernoulli}(z_{ik} ; \pi_k)
\end{equation}
which leads to an equivalent Beta distribution as 
\begin{equation}
    p(\pi_k \mid -) \sim \text{Beta} \left( \frac{a}{K} + \sum_{i=1}^{n_{total}} z_{ik}, \; \frac{b(K-1)}{K} + {n_{total}} - \sum_{i=1}^{n_{total}} z_{ik} \right)
\end{equation}
The hyperparameter \(\gamma_s\) associated with the distribution of coefficient vector \(\mathbf{s_i}\) can be drawn from a Gamma PDF given by
\begin{equation}
p(\gamma_s \mid -) \sim \Gamma \left( 
c + \frac{1}{2} K {n_{total}}, \;
d + \frac{1}{2} \sum_{i=1}^{n_{total}} \mathbf{s}_i^\top \mathbf{s}_i 
\right)
\end{equation}
whereas the distribution of the \(\gamma_\epsilon\) hyperparameter related to the error \(\mathbf{\epsilon}_i\) is
\begin{equation}
    p(\gamma_\epsilon \mid -) \propto \Gamma(\gamma_\epsilon ; e, f) \prod_{i=1}^{{n_{total}}} \mathcal{N}\left( \mathbf{y}_i ; \mathbf{\Sigma}_i \mathbf{D}(\mathbf{s}_i \odot \mathbf{z}_i), \gamma_\epsilon^{-1} \mathbf{I}_{\|\mathbf{\Sigma}_i\|_0} \right)
\end{equation}
It can be shown \cite{zhou2011nonparametric} that this leads to the following Gamma distribution, i.e.,
\begin{equation}
    p(\gamma_\epsilon \mid -) \sim \Gamma \left( 
    e + \frac{1}{2} \sum_{i=1}^{n_{total}} \|\mathbf{\Sigma}_i\|_0, \;
    f + \frac{1}{2} \sum_{i=1}^{n_{total}} 
    \left\| 
    \mathbf{\Sigma}_i^\top \mathbf{y}_i - \mathbf{\Sigma}_i^\top \mathbf{\Sigma}_i \mathbf{D}(\mathbf{s}_i \odot \mathbf{z}_i) 
    \right\|_{\ell_2}^2 
    \right)
\end{equation}

\bibliographystyle{elsarticle-num}
\biboptions{sort&compress}
\bibliography{wind_DL_UQ}

 \end{document}